\documentclass[11pt,leqno,oneside]{amsart}  
\usepackage{graphicx}
\usepackage{indentfirst,csquotes}

\usepackage{amssymb,amsthm,amsmath}
\usepackage{paralist,hyperref,titlesec,fancyhdr,etoolbox}

\usepackage{titlesec}

\titleformat{\section}[hang]
  {\normalfont\LARGE\bfseries}{\thesection}{1em}{}

\titleformat{\subsection}[hang]
  {\normalfont\Large\bfseries}{\thesubsection}{1em}{}

\titleformat{\subsubsection}[hang]
  {\normalfont\normalsize\bfseries}{\thesubsubsection}{1em}{}

\titlespacing*{\section}{0pt}{1.5ex plus 1ex minus .2ex}{1ex}
\titlespacing*{\subsection}{0pt}{1.25ex plus 1ex minus .2ex}{1ex}
\titlespacing*{\subsubsection}{0pt}{1ex plus .5ex minus .2ex}{1ex}

\hypersetup{ colorlinks=true, linkcolor=black, filecolor=black, urlcolor=black }

\usepackage{lipsum}
\usepackage{float}
\usepackage{setspace} 
\usepackage{amssymb}
\usepackage{amsmath}
\usepackage{booktabs}
\usepackage{amssymb,amsmath,amsfonts,amscd,verbatim}
\usepackage{url,upgreek}
\usepackage{graphicx}
\usepackage{mathrsfs}
\usepackage{amssymb}
\usepackage{amsmath}
\usepackage{bbm}
\usepackage{url,upgreek}
\usepackage{float}
\usepackage[font={small}]{caption}
\usepackage{geometry}
\usepackage{setspace}
\usepackage{amssymb,amsmath,amsfonts,amscd,verbatim}
\usepackage{url,upgreek}
\usepackage{amsmath,amsfonts,amsthm,amssymb}
\usepackage{setspace}
\usepackage{Tabbing}
\usepackage{fancyhdr}
\usepackage{lastpage}
\usepackage{extramarks}
\usepackage{chngpage}
\usepackage{soul,color}
\usepackage{graphicx,float,wrapfig}
\usepackage{amsfonts}
\usepackage{bbold}
\usepackage{soul}
\newcommand\textopenone{\leavevmode\hbox{\small 1\kern-3.3pt\normalsize 1}}

\DeclareMathOperator{\rank}{\mathrm{rank}}

\newcommand{\real}{\mathbb{R}}

\newcommand\vb{\mathbf{b}}

\newcommand\vg{\mathbf{g}}

\newcommand\vv{\mathbf{v}}

\newcommand\mg{\mathbf{G}}
\newcommand\mh{\mathbf{H}}

\newcommand\ms{\mathbf{S}}

\newcommand\mv{\mathbf{V}}
\newcommand\mw{\mathbf{W}}

\newcommand\mz{\mathbf{Z}}
\usepackage{bm}
\usepackage{amssymb,amsmath,amsfonts,amscd,verbatim}
\usepackage{url,upgreek}
\usepackage{graphicx}
\usepackage{mathrsfs}
\usepackage{amssymb,amsmath,amsfonts,amscd,verbatim}
\usepackage{url,upgreek}
\usepackage{amsmath,amsfonts,amsthm,amssymb}
\usepackage{setspace}
\usepackage{Tabbing}
\usepackage{fancyhdr}
\usepackage{lastpage}
\usepackage{extramarks}
\usepackage{chngpage}
\usepackage{soul,color}
\usepackage{graphicx,float,wrapfig}
\usepackage{amsfonts}
\usepackage{bbold}
\usepackage{hyperref}
\usepackage{mathtools}
\usepackage{bbm}
\usepackage{url,upgreek}
\usepackage{float}
\usepackage[font={small}]{caption}
\usepackage{geometry}
\usepackage{listings}
\usepackage{setspace}
\usepackage[table,xcdraw,dvipsnames]{xcolor}
\usepackage{color,soul}
\usepackage[linesnumbered,ruled,lined]{algorithm2e}
\SetKwInput{KwInput}{Input}                
\SetKwInput{KwOutput}{Output}              
\newcommand\vbeta{\boldsymbol{\beta}}

\newcommand{\combn}{\boldsymbol{x}_{ij}^\top\boldsymbol{\beta}}
\newcommand{\lp}{\Tilde{\ell}_i^{\prime}}
\newcommand{\tbeta}{\tilde{\boldsymbol{\beta}}}
\newcommand{\tsigma}{\Tilde{\sigma}}

\newcommand{\tcombn}{\boldsymbol{x}_{ij}^\top\Tilde{\boldsymbol{\beta}}}

\newcommand{\tx}{\boldsymbol{x}_{ij}}
\DeclareMathOperator{\Null}{\mathrm{Null}}

\usepackage{tikz}

\newtheorem{Proposition}{Proposition}
\DeclarePairedDelimiter\ceil{\lceil}{\rceil}
\DeclarePairedDelimiter\floor{\lfloor}{\rfloor}

\usepackage{siunitx,booktabs,threeparttable,caption}

\usepackage{bbm}

\begin{document}
\title{A Two-Stage Federated Learning Approach for Industrial Prognostics Using Large-Scale High-Dimensional Signals}

\author{Yuqi Su$^1$}
\author{Xiaolei Fang$^2$}

\let\thefootnote\relax
\footnotetext{This work has been submitted to the IEEE for possible publication. Copyright may be transferred without notice, after which this version may no longer be accessible.}
\footnotetext{\textsuperscript{1}Operations Research Graduate Program, North Carolina State University, Raleigh, NC, USA.  Email: ysu25@ncsu.edu}
\footnotetext{\textsuperscript{2}Edward P. Fitts Department of Industrial and Systems Engineering, North Carolina State University, Raleigh, NC, USA. Email: xfang8@ncsu.edu}

\begin{abstract}
Industrial prognostics aims to develop data-driven methods that leverage high-dimensional degradation signals from assets to predict their failure times. The success of these models largely depends on the availability of substantial historical data for training. However, in practice, individual organizations often lack sufficient data to independently train reliable prognostic models, and privacy concerns prevent data sharing between organizations for collaborative model training. To overcome these challenges, this article proposes a statistical learning-based federated model that enables multiple organizations to jointly train a prognostic model while keeping their data local and secure. The proposed approach involves two key stages: federated dimension reduction and federated (log)-location-scale regression. In the first stage, we develop a federated randomized singular value decomposition algorithm for multivariate functional principal component analysis, which efficiently reduces the dimensionality of degradation signals while maintaining data privacy. The second stage proposes a federated parameter estimation algorithm for (log)-location-scale regression, allowing organizations to collaboratively estimate failure time distributions without sharing raw data. The proposed approach addresses the limitations of existing federated prognostic methods by using statistical learning techniques that perform well with smaller datasets and provide comprehensive failure time distributions. The effectiveness and practicality of the proposed model are validated using simulated data and a dataset from the NASA repository.
\end{abstract}
\maketitle

\bigskip

\section{Introduction}\label{section1}
Critical engineering assets are typically equipped with numerous sensors to monitor their health condition. Some of these sensors are able to capture the underlying degradation process, which is known as the process of damage accumulation that eventually leads to asset failure. The signals from these sensors are commonly referred to as degradation signals. When properly modeled, these signals can be highly valuable for industrial prognostics, which aims to predict the failure times of critical engineering assets using their real-time degradation data. Industrial prognostics is typically achieved by training statistical learning or machine learning models that map an asset's degradation signals to its time-to-failure (TTF). 

Similar to other data-driven models, prognostic models usually require a sufficient amount of historical data for model training. However, in practice, data are often distributed across various organizations such as companies, factories, original equipment manufacturers (OEMs), and college labs. Despite the daily generation of large volumes of condition monitoring data, the records of failures from any single organization are typically quite limited. Consequently, it can be challenging for a single organization to independently train a high-performance prognostic model. Usually, organizations can benefit from collaboratively training a prognostic model with the union of their data. 

One potential technique that allows multiple organizations to jointly train a prognostic model is \textit{cloud computing}. This approach works by allowing each organization to upload its data to a server on the cloud. The server then aggregates the data from all organizations and uses them for prognostic model training. The trained model is then sent back to each organization for real-time condition monitoring and failure time prediction. While cloud computing facilitates collaborative model construction by pooling data resources, it raises a significant challenge related to \textit{privacy protection}. Due to various privacy protection policies and regulations, data from different organizations cannot be simply shared or merged. Many organizations are not even allowed to upload their data to the cloud \cite{pardau2018california,voigt2017eu,chik2013singapore}. Consequently, data in different organizations are usually {{isolated}} and become ``data islands."

An effective technique that can address the privacy concern is \textit{federated learning} (FL) \cite{konevcny2016federated}. FL allows multiple organizations to collaboratively train machine learning models using their isolated data. This technique ensures that each organization's data remains local and confidential, effectively overcoming the barriers imposed by traditional cloud computing approaches. Although there has been some progress in integrating federated learning into industrial prognostics \cite{guo2022fedrul, chen2023bearing, cai2024fedcov, arunan2023federated, kamei2023comparison, zhu2024collaborative}, these efforts face several common challenges. First, most federated prognostic models are based on deep learning, with few leveraging statistical learning models. Although deep learning is effective in various contexts, statistical learning often outperforms deep learning when dealing with smaller sample sizes---a common scenario in industrial prognostic applications, even when data from multiple organizations are combined. Second, existing federated prognostic models generally provide only point estimates of the predicted failure time. However, many decision-making tasks in equipment health management, such as maintenance, logistics, and inventory optimization, require a distribution of predicted failure times rather than just a single point estimate.  This gap makes it challenging to fully integrate these models into practical decision-making processes. 

To address these challenges, this article proposes a statistical learning-based federated prognostic model. The proposed model comprises two key stages: \textit{federated dimension reduction} and \textit{federated (log)-location-scale (LLS) Regression}. In the first stage, \textit{federated dimension reduction}, we employ multivariarte functional principal component analysis (MFPCA) to fuse multi-stream high-dimensional degradation signals. MFPCA is effective in prognostic applications involving multi-sensor degradation signals \cite{fang2017scalable, fang2021multi} due to its ability to capture both the auto-correlation within each sensor and the cross-correlation across multiple sensors. It reduces the dimension of degradation signals and provides low-dimensional features known as MFPC-scores, which can be used to construct a prognostic model in the second stage (i.e., \textit{federated LLS regression}). The most commonly used method to calculate MFPC-scores is singular value decomposition (SVD). However, traditional SVD requires a centered dataset, meaning that data from all users must be aggregated first, making it unsuitable for federated learning models. While some existing works have developed federated SVD algorithms that enable multiple users to perform SVD collaboratively while keeping their data local and confidential \cite{chai2022practical, hartebrodt2024federated, liu2023privacy}, these algorithms face \textit{high communication and computation costs}. This is largely because these methods achieve privacy protection by adding masking information to the raw data and sending the masked data to the central server. In prognostic applications, the volume of data sent to the server is substantial. This is because although the number of failure records (e.g., training samples) from an organization is typically small, the amount of sensing data associated with each record is often enormous, as many assets are monitored by numerous sensors generating vast amounts of degradation data. The cost of uploading the masked ``Big Data" to the cloud and analyzing it can be prohibitively high.

To address these challenge, this article develops a \textit{federated randomized singular value decomposition (FRSVD)} algorithm. RSVD is an efficient and scalable algorithm for approximating the SVD of large scale matrices. RSVD is particularly useful when dealing with high-dimensional data or large matrices, where traditional SVD methods can be computationally expensive and memory-intensive \cite{halko2011finding}. RSVD works by first identifying the low-dimensional subspace spanned by the dominant singular vectors, then projecting the original data onto this subspace and performing SVD within it. This approach significantly reduces computational effort because the SVD is conducted in a much lower-dimensional space. Leveraging this advantage, the proposed FRSVD algorithm not only lowers computational costs but also substantially reduces communication costs, as the data exchanged between users and the server are significantly lower-dimensional compared to the original data.

The second stage of the proposed statistical learning-based federated prognostic model, \textit{federated LLS regression}, focuses on constructing a prediction function that maps the asset's features extracted in the first stage (i.e., MFPC-scores) to its time-to-failure. This is achieved using (log)-location-scale regression, a widely used method in reliability and survival analysis \cite{william1998statistical} due to its flexibility in modeling various failure time distributions, such as (log)-normal, (log)-logistic, smallest extreme value, and Weibull distributions. A key advantage of LLS regression is its ability to provide a distribution of the predicted failure time, which is crucial for decision-making tasks such as maintenance, logistics, and inventory optimization. Parameter estimation for LLS regression is commonly performed using maximum likelihood estimation (MLE). However, existing optimization algorithms for solving MLE in LLS regression often require the data to be centered, making it incompatible with federated learning. To address this challenge, this article develops a \textit{federated parameter estimation algorithm} that allows multiple users to jointly estimate the parameters of the LLS regression while keeping their data local and confidential. The proposed federated parameter estimation algorithm is based on gradient descent, an iterative optimization method that optimizes an objective function by repeatedly moving in the direction of the steepest descent. We have observed that the total gradient in each iteration is the sum of the gradients from all individual training samples. This insight allows each user to compute their local gradient using local data and share these local gradients with the server on the cloud. The server then aggregates the gradients of all users to update the model parameters. Since only the local gradients---not the raw data (i.e., MFPC-scores and TTFs)---are exchanged between the server and users, data privacy is effectively protected.

The rest of the article is organized as follows. Sections \ref{sec:fdr} and \ref{sec:LLS} provide detailed discussions of the proposed federated dimension reduction method and the federated LLS regression model, respectively. In Section \ref{section3.5}, we evaluate the effectiveness of the proposed federated prognostics model using simulated data, followed by an assessment using a dataset from the NASA data repository in Section \ref{section4}. Finally, Section \ref{section5} presents the conclusions.

\section{Federated Dimension Reduction}\label{sec:fdr}
\subsection{Multivariate Functional Principal Component Analysis}\label{sec:mpca}

Let $I$ denote the number of users participating in the development of the federated prognostic model. Let $J_i$  be the number of failed assets in the local training dataset of user $i$. Moreover, suppose each asset is characterized by $P$-channel degradation signals, where ${s}_{ij}^{(p)}(t)$ represents the degradation signal from sensor $p$ of system $j$ belonging to user $i$, with $i=1,\ldots, I$, $j=1,\ldots, J_i$, and $p=1,\ldots, P$. Here, $t \in [0,T]$, where $[0, T]$ denotes the degradation time interval. We use $\boldsymbol{s}_{ij}(t)=(s_{ij}^{(1)}(t),s_{ij}^{(2)}(t),...,s_{ij}^{(P)}(t))^\top$ to denote the concatenated degradation signals from all $P$ sensors of system $j$ of user $i$. $\{\{\boldsymbol{s}_{ij}(t)\}_{j=1}^{J_i}\}_{i=1}^{I}$ can be seen as independent realizations of a $p$-dimensional stochastic process with mean vector $\boldsymbol{\mu}(t)=(\mu^{(1)}(t), \mu^{(2)}(t),...,\mu^{(P)}(t))^\top$ and a $P \times P$ block covariance matrix $\boldsymbol{C}(t,t')$, where the $(g,h)$th entry, ${C}_{g,h}(t,t')$, is the covariance function between sensor $g$ and $h$, $g=1,\ldots,P, h=1,\ldots,P$. 

Based on the Karhunen-Lo\`eve theorem \cite{karhunen1947ueber}, the concatenated degradation signal $\boldsymbol{s}_{ij}(t)$ can be decomposed as follows:
\begin{equation}\label{eq3}
    \boldsymbol{s}_{ij}(t) = \boldsymbol{\mu}(t) + \sum_{k=1}^\infty{x}_{ijk}\boldsymbol{\phi}_k(t), 
\end{equation}

\noindent where $\boldsymbol{\phi}_k(t) = (\phi_{k}^{(1)}(t),\phi_{k}^{(2)}(t),...,\phi_{k}^{(P)}(t))^\top$ is the $k$th eigen function of $\boldsymbol{C}(t,t')$, ${x}_{ijk}=\int_0^\top (\boldsymbol{s}_{ij}(t)-\boldsymbol{\mu}(t))^\top\boldsymbol{\phi}_k(t)dt$ is the feature known as MFPC-score. Also, it is known that the mean of ${x}_{ijk}$, $\mathbb{E}({x}_{ijk})=0$, variance $\mathbb{V}({x}_{ijk})=\lambda_k$, and the covariance $Cov({x}_{ijk},{x}_{ijl})=0$ if $k\neq l$, where $\lambda_k$ is the $k$th eigenvalue of $\boldsymbol{C}(t,t')$. Usually, the eigenvalues $\lambda_1\geq\lambda_2\geq\ldots$ decrease rapidly. Thus, 
 it is sufficient to keep the first $K$ eigenfunctions corresponding to the $K$ largest eigenvalues where $K$ can be determined using Fraction-of-Variance Explained (FVE) \cite{fang2015adaptive} or cross-validation \cite{fang2017multistream}. As a result, the  degradation signal $\boldsymbol{s}_{ij}(t)$ can be approximated as follows: $\boldsymbol{s}_{ij}(t) \approx \boldsymbol{\mu}(t) +\sum_{k=1}^K{x}_{ijk}\boldsymbol{\phi}_k(t)$, which implies that the degradation signal $\boldsymbol{s}_{ij}(t)$ can be represented using MFPC-scores $\{{x}_{ijk}\}_{k=1}^K$. This is because all signals $\{\{\boldsymbol{s}_{ij}(t)\}_{j=1}^{J_i}\}_{i=1}^{I}$ share the same mean function $\boldsymbol{\mu}(t)$ and basis function $\{\boldsymbol{\phi}_k(t)\}_{k=1}^K$. What distinguishes them are the MFPC-scores, which capture the unique characteristics of each user's degradation data. Thus, instead of employing the high-dimensional signal $\boldsymbol{s}_{ij}(t)$ to construct a prognostic model, it is common to utilize its low-dimensional features $\{{x}_{ijk}\}_{k=1}^K$ \cite{fang2017multistream,fang2017scalable,fang2021multi}.

Recall that ${x}_{ijk}=\int_0^\top (\boldsymbol{s}_{ij}(t)-\boldsymbol{\mu}(t))^\top\boldsymbol{\phi}_k(t)dt$. Solving this integral is challenging not only due to the complexity of the integral itself, but also because the observed degradation signals are discrete rather than continuous. As a result, $\{{x}_{ijk}\}_{k=1}^K$ are usually computed by using discrete techniques \cite{yao2005functional}. Specifically, if we denote the discrete observation time as $\tau_1,\tau_2,\ldots,\tau_m$, then the degradation signal from sensor $p$ of the $j$th asset of user $i$ can be denoted as $\boldsymbol{s}_{ij}^{(p)}=((s_{ij}^{(p)}(\tau_1)),s_{ij}^{(p)}(\tau_2),\ldots,s_{ij}^{(p)}(\tau_m))^\top\in\mathbb{R}^{m}$. Thus, the concatenated signal from all $P$ sensors of  the $j$th asset of user $i$ can be expressed as $\boldsymbol{s}_{ij}=(\boldsymbol{s}_{ij}^{(1)\top},\boldsymbol{s}_{ij}^{(2)\top},\ldots,\boldsymbol{s}_{ij}^{(P)\top})^\top\in\mathbb{R}^{M}$, where $M=m\times P$. The first method that can be employed to compute ${x}_{ijk}$ is eigendecomposition (ED). Specifically, we first calculate the mean vector $\bar{\boldsymbol{\mu}}=(\sum_{i=1}^{I}\sum_{j=1}^{J_i}\boldsymbol{s}_{ij})/(\sum_{i=1}^IJ_i)^\top\in\mathbb{R}^{M}$. Then, derive the covariance matrix $\boldsymbol{C}=\sum_{i=1}^{I}\sum_{j=1}^{J_i}(\boldsymbol{s}_{ij}-\bar{\boldsymbol{\mu}})(\boldsymbol{s}_{ij}-\bar{\boldsymbol{\mu}})^\top\in\mathbb{R}^{M\times M}$ across all data samples. Next, we perform eigendecomposition on $\boldsymbol{C}$, i.e., $\boldsymbol{C}\boldsymbol{v}_k=\lambda_k\boldsymbol{v}_k$, to obtain eigenvectors $\boldsymbol{v}_k\in\mathbb{R}^{M}$ and eigenvalues $\lambda_k$ for $k = 1, \ldots, K$. Finally, compute ${x}_{ijk}=(\boldsymbol{s}_{ij}-\bar{\boldsymbol{\mu}})^\top\boldsymbol{v}_k$. Another way to compute ${x}_{ijk}$ is to perform SVD on the centered signal matrix containing the discrete observations from all assets. Specifically, let $\boldsymbol{S}=\left(\boldsymbol{s}_{11},\ldots,\boldsymbol{s}_{1J_1},\boldsymbol{s}_{21},\ldots,\boldsymbol{s}_{2J_2}\ldots\boldsymbol{s}_{I1},\ldots,\boldsymbol{s}_{IJ_I}\right)^\top\in\mathbb{R}^{J\times M}$ and $\tilde{\boldsymbol{S}}$ be the centered $\boldsymbol{S}$ by reducing each of its row by $\bar{\boldsymbol{\mu}}$, where $J=\sum_{i=1}^IJ_i$ is the total number of samples that all the users have. Then, $\tilde{\boldsymbol{S}}=\boldsymbol{U}\boldsymbol{\Sigma}\boldsymbol{V}^\top$, where the columns of $\boldsymbol{U}$ are known as the left singular vectors, and the columns of $\boldsymbol{V}$ are known as the right singular vectors. The first $K$ columns of $\boldsymbol{V}$ are the same as the $K$ eigenvectors of the covariance matrix $\boldsymbol{C}$ (i.e., $\{\boldsymbol{v}_k\}_{k=1}^K$). 

It can be seen that computing MFPC-scores ${x}_{ijk}$ using classic ED or SVD requires aggregating data from all users, which compromises data privacy. To address this challenge, a federated learning algorithm needs to be developed. While some existing federated methods in the literature can be employed to compute MFPC-scores, they often suffer from high communication and computation costs. To overcome these issues, this paper proposes a Federated Randomized Singular Value Decomposition (FRSVD) algorithm for MFPC-scores computation, the details of which are provided in the subsequent Section \ref{sec:frsvd}.

\subsection{Federated Randomized Singular Value Decomposition}\label{sec:frsvd}

Randomized Singular Value Decomposition (RSVD) is a technique for efficiently approximating the SVD of large matrices, as discussed in \cite{halko2011finding}. This method is particularly valuable in scenarios involving very large datasets where traditional SVD would be computationally prohibitive. RSVD offers a more efficient alternative for computing MFPC-scores compared to classic SVD. In classic SVD, all singular vectors are calculated initially; however, only the first $K$ right singular vectors---corresponding to the first $K$ principal components---are retained, with the rest being discarded. Typically, $K$ represents a small fraction of the total singular vectors, meaning that most of the computation effort expended on calculating unused singular vectors is wasted, resulting in unnecessary consumption of computational resources and time. Unlike SVD, RSVD targets this inefficiency by directly computing only the first $K$ singular vectors. It accomplishes this by identifying the subspace spanned by these vectors and then performing classic SVD within that subspace. More technical details of RSVD will be explored in the following paragraphs.

Let $\tilde{\boldsymbol{S}}\in\mathbb{R}^{J\times M}$ be the centered signal matrix, where $J$ is the number of samples, and $M$ is the signal length of each sample. Let the rank of $\tilde{\boldsymbol{S}}$ be $R$, $R\leq \min\{J,M\}$, then the SVD of $\tilde{\boldsymbol{S}}$ is $\tilde{\boldsymbol{S}}=\sum_{i=1}^R\sigma_i\boldsymbol{u}_i\boldsymbol{v}_i^\top$, where $\{\sigma_i\}_{i=1}^R$, $\{\boldsymbol{u}_i\}_{i=1}^R$, and $\{\boldsymbol{v}_i\}_{i=1}^R$ are the singular values, left singular vectors, and right singular vectors, respectively. Recall that $K$ is the number of principal components, then we can split $\tilde{\boldsymbol{S}}$ into two components: $\tilde{\boldsymbol{S}}=\tilde{\boldsymbol{S}}_K+\tilde{\boldsymbol{S}}_{-K}=\sum_{i=1}^K\sigma_i\boldsymbol{u}_i\boldsymbol{v}_i^\top+\sum_{i=K+1}^R\sigma_i\boldsymbol{u}_i\boldsymbol{v}_i^\top$, where $\tilde{\boldsymbol{S}}_K=\sum_{i=1}^K\sigma_i\boldsymbol{u}_i\boldsymbol{v}_i^\top$ is constructed from the first $K$ singular values and vectors, and $\tilde{\boldsymbol{S}}_{-K}=\sum_{i=K+1}^R\sigma_i\boldsymbol{u}_i\boldsymbol{v}_i^\top$ is constructed from the remaining singular values and vectors. Keep in mind that the objective of RSVD is to compute the singular values and vectors in $\tilde{\boldsymbol{S}}_K$ while avoiding the computation of those in $\tilde{\boldsymbol{S}}_{-K}$.

The \textit{first} step of RSVD is to identify the subspace spanned by the columns of $\tilde{\boldsymbol{S}}_K$ using randomized sampling. This is achieved by multiplying $\Tilde{\boldsymbol{S}}$ by a Gaussian random matrix $\mw\in\real^{M\times K}$. It is known that any linear combination of the columns of a matrix lies in the range (i.e., the column space) of the matrix, so $\{\tilde{\boldsymbol{S}}\boldsymbol{w}_k\}_{k=1}^K$ lie in the range of $\Tilde{\boldsymbol{S}}$, where $\boldsymbol{w}_k$ is the $k$th column of $\mw$. Also, since vectors $\{\boldsymbol{w}_k\}_{k=1}^K$ are randomly generated, they are almost surely in a general position such that no linear combination of
these vectors falls into the null space of $\Tilde{\boldsymbol{S}}$. As a result, $\{\tilde{\boldsymbol{S}}\boldsymbol{w}_k\}_{k=1}^K$ are almost surely linearly independent. Based on the rank of $\tilde{\boldsymbol{S}}$, we can identify the following two scenarios: (i) If the rank of $\tilde{\boldsymbol{S}}$ is $K$ (that is, $R=K$, $\tilde{\boldsymbol{S}}_{-K}$ does not exist, and $\tilde{\boldsymbol{S}}=\tilde{\boldsymbol{S}}_K$), then $\{\tilde{\boldsymbol{S}}\boldsymbol{w}_k\}_{k=1}^K$ span the range of $\tilde{\boldsymbol{S}}_{K}$ (or $\tilde{\boldsymbol{S}}$). In practice, it is possible (although highly unlikely) that the vectors $\{\tilde{\boldsymbol{S}}\boldsymbol{w}_k\}_{k=1}^K$ are not linearly independent. To address this issue, we can typically sample $K+r$ instead of $K$ random vectors (i.e., $\{\boldsymbol{w}_k\}_{k=1}^{K+r}$), where $r$ is a small oversampling number. (ii) If the rank of $\tilde{\boldsymbol{S}}$ is larger than $K$, then $\{\tilde{\boldsymbol{S}}\boldsymbol{w}_k\}_{k=1}^K$ approximately span the range of $\tilde{\boldsymbol{S}}_K$. It is an approximation because $\tilde{\boldsymbol{S}}\boldsymbol{w}_k=\tilde{\boldsymbol{S}}_K\boldsymbol{w}_k+\tilde{\boldsymbol{S}}_{-K}\boldsymbol{w}_k$, where $\tilde{\boldsymbol{S}}_{-K}\boldsymbol{w}_k$ can be seen as a perturbation that shifts the vector $\tilde{\boldsymbol{S}}\boldsymbol{w}_k$ outside the range of $\tilde{\boldsymbol{S}}_K$. To minimize the impact of perturbation, we may use $\{(\tilde{\boldsymbol{S}}\tilde{\boldsymbol{S}}^\top)^q\tilde{\boldsymbol{S}}\boldsymbol{w}_k\}_{k=1}^K$ instead of $\{\tilde{\boldsymbol{S}}\boldsymbol{w}_k\}_{k=1}^K$, where $q$ is a small number such as 1 or 2 \cite{halko2011finding}.  This approach is effective because $(\tilde{\boldsymbol{S}}\tilde{\boldsymbol{S}}^\top)^q\tilde{\boldsymbol{S}}\boldsymbol{w}_k=\sum_{i=1}^K\sigma_i^{2q+1}\boldsymbol{u}_i\boldsymbol{v}_i^\top\boldsymbol{w}_k+\sum_{i=K+1}^R\sigma_i^{2q+1}\boldsymbol{u}_i\boldsymbol{v}_i^\top\boldsymbol{w}_k$. Since $\{\sigma_i\}_{i=K+1}^{R}$ are much smaller than $\{\sigma_i\}_{i=1}^{K}$, the exponent term $2q+1$ further amplifies the difference between the two groups. In fact, $\{\sigma_i\}_{i=K+1}^{R}$ are often much smaller than 1, so $\{\sigma_i^{2q+1}\}_{i=K+1}^{R}$ are nearly zeros. Therefore, the perturbation in $(\tilde{\boldsymbol{S}}\tilde{\boldsymbol{S}}^\top)^q\tilde{\boldsymbol{S}}\boldsymbol{w}_k$ is negligible. In other words, the vectors $\{(\tilde{\boldsymbol{S}}\tilde{\boldsymbol{S}}^\top)^q\tilde{\boldsymbol{S}}\boldsymbol{w}_k\}_{k=1}^K$ approximately span the range of $\sum_{i=1}^K\sigma_i^{2q+1}\boldsymbol{u}_i\boldsymbol{v}_i^\top$. Since $\sum_{i=1}^K\sigma_i^{2q+1}\boldsymbol{u}_i\boldsymbol{v}_i^\top$ and $\tilde{\boldsymbol{S}}_K$ share the same range, the vectors $\{(\tilde{\boldsymbol{S}}\tilde{\boldsymbol{S}}^\top)^q\tilde{\boldsymbol{S}}\boldsymbol{w}_k\}_{k=1}^K$ approximately span the range of $\tilde{\boldsymbol{S}}_K$. Similar to scenario (i), it is customary to sample $r$ additional vectors to mitigate the unlikely event of linear dependence among the vectors $\{(\tilde{\boldsymbol{S}}\tilde{\boldsymbol{S}}^\top)^q\tilde{\boldsymbol{S}}\boldsymbol{w}_k\}_{k=1}^K$. This results in the extended set $\{(\tilde{\boldsymbol{S}}\tilde{\boldsymbol{S}}^\top)^q\tilde{\boldsymbol{S}}\boldsymbol{w}_k\}_{k=1}^{K+r}$ or in matrix form, $(\tilde{\boldsymbol{S}}\tilde{\boldsymbol{S}}^\top)^q\tilde{\boldsymbol{S}}\boldsymbol{W}$, where $\boldsymbol{W}\in\real^{M\times(K+r)}$. 

The \textit{second} step of RSVD is to extract a set of orthonormal basis vectors for the range of $\tilde{\boldsymbol{S}}_K$. This can be done by performing QR decomposition on the matrix $\tilde{\boldsymbol{S}}\boldsymbol{W}$ (when the rank of $\tilde{\boldsymbol{S}}$ equals $K$) or on $(\tilde{\boldsymbol{S}}\tilde{\boldsymbol{S}}^\top)^q\tilde{\boldsymbol{S}}\boldsymbol{W}$ (when the rank of $\tilde{\boldsymbol{S}}$ is larger than $K$). The resulting basis matrix is denoted by $\boldsymbol{Q}\in\real^{J\times K}$. In the \textit{third} step, a low-dimensional matrix $\boldsymbol{B}=\boldsymbol{Q}^\top\tilde{\boldsymbol{S}}\in\real^{K\times L}$ is computed. Then, compact SVD is performed on $\boldsymbol{B}$, yielding $\boldsymbol{B}=\hat{\boldsymbol{U}}\hat{\boldsymbol{\Sigma}}\hat{\boldsymbol{V}}^\top$. Recall that we have denoted the SVD of $\tilde{\boldsymbol{S}}$ as $\tilde{\boldsymbol{S}}=\boldsymbol{U}\boldsymbol{\Sigma}\boldsymbol{V}^\top$, then it can be shown that $\boldsymbol{U}$, $\boldsymbol{\Sigma}$, and $\boldsymbol{V}$ can be computed as follows: $\boldsymbol{U}=\boldsymbol{Q}\hat{\boldsymbol{U}}\in\real^{J\times K}$, $\boldsymbol{\Sigma}=\hat{\boldsymbol{\Sigma}}\in\real^{K\times K}$, and $\boldsymbol{V}=\hat{\boldsymbol{V}}\in\real^{L\times K}$ \cite{fang2017scalable}.

Based on RSVD, this article develops a federated randomized singular value decomposition (FRSVD) algorithm that can be used to compute the MFPC-scores. This development is nontrivial due to several reasons. First, the raw signal matrix (i.e., $\boldsymbol{S}$) needs to be centered to produce $\tilde{\boldsymbol{S}}$ before performing RSVD. This preprocessing step is a requirement of MFPCA but not RSVD. To center the data in the federated learning setting, one possible solution is to ask each user to compute a local mean from their data and then send this mean along with their sample size to the server. The server then calculates a global mean through a weighted average of these local means. Subsequently, each user downloads the global mean from the server and centers their local data accordingly. However, one issue with this solution is that the local means themselves can be confidential and cannot be shared with the server. To address this issue, we could perform a slightly revised RSVD on the uncentered signal matrix $\boldsymbol{S}$. Specifically, we modify the third step of RSVD by centering the low-dimensional matrix $\boldsymbol{B}$ before conducting SVD on it. In an earlier work \cite{fang2021multi}, the authors have demonstrated that the singular values and vectors obtained by this method are equivalent to those derived from performing RSVD on the centered matrix $\tilde{\boldsymbol{S}}$. Therefore, in the proposed FRSVD, we will use $\boldsymbol{S}$ instead of $\tilde{\boldsymbol{S}}$.

Another, and yet more significant, challenge when developing the FRSVD algorithm is how to enable multiple users to jointly perform RSVD while keeping each user's data local and confidential. In the following paragraphs, we will outline the challenges associated with each of the three steps of RSVD and present our proposed solutions. 

In the \textit{first} step of RSVD, the subspace is identified through randomized sampling. This requires computing $\boldsymbol{S}\boldsymbol{W}$ when the rank of $\boldsymbol{S}$ equals $K$, or $(\boldsymbol{S}\boldsymbol{S}^\top)^q\boldsymbol{S}\boldsymbol{W}$ when the rank of $\boldsymbol{S}$ exceeds $K$. However, $\boldsymbol{S}$ is not available since the first $J_1$ rows of $\boldsymbol{S}$ belong to user 1, and the next $J_2$ rows belong to user 2, etc. In other words, $\boldsymbol{S}=(\boldsymbol{S}_1^\top, \boldsymbol{S}_2^\top,\ldots,\boldsymbol{S}_I^\top)^\top$, where $\boldsymbol{S}_i\in\real^{J_i\times L}$ is the local signal matrix of user $i$. To compute $\boldsymbol{S}\boldsymbol{W}$, we notice that $\boldsymbol{S}\boldsymbol{W}=((\boldsymbol{S}_1\boldsymbol{W})^\top, (\boldsymbol{S}_2\boldsymbol{W})^\top,\ldots,(\boldsymbol{S}_I\boldsymbol{W})^\top)^\top$. This inspires us to (i) let the server generate the random matrix $\boldsymbol{W}\in\mathbb{R}^{L\times (K+r)}$, (ii) each user $i$ then downloads $\boldsymbol{W}$ from the server, locally computes $\boldsymbol{Y}_i=\boldsymbol{S}_i\boldsymbol{W}\in\mathbb{R}^{J_i\times (K+r)}$, and sends $\boldsymbol{Y}_i$ to the server, and (iii) the server aggregates $\boldsymbol{Y}=(\boldsymbol{Y}_1^\top,\boldsymbol{Y}_2^\top,\ldots, \boldsymbol{Y}_I^\top)^\top\in\mathbb{R}^{J\times (K+r)}$, where $J=\sum_{i=1}^IJ_i$ is the total number of samples from all users. Although this allows multiple users to jointly compute $\boldsymbol{Y}$ (i.e., $\boldsymbol{S}\boldsymbol{W}$), a crucial question arises: Is the server able to recover the data of user $i$ from $\boldsymbol{Y}_i$? In other words, given $\boldsymbol{W}$ and $\boldsymbol{Y}_i=\boldsymbol{S}_i\boldsymbol{W}$, can $\boldsymbol{S}_i$ be recovered? To answer this question, we provide the following Proposition. 

\begin{Proposition}\label{prop:recovery}
Given $\mz=\ms\mg$, where $\mz$ and $\mg$ are known, then $\ms$ cannot be uniquely recovered if $\mg$ is not full row rank.
\end{Proposition}

The proof of Proposition \ref{prop:recovery} is detailed in Section \ref{proof1}. Recall that $\boldsymbol{Y}_i \in \mathbb{R}^{J_i \times (K+r)}$, $\boldsymbol{S}_i \in \mathbb{R}^{J_i \times L}$, and $\boldsymbol{W} \in \mathbb{R}^{L \times (K+r)}$, where $L$ represents the length of the signals, $K$ denotes the number of principal components, and $r$ is the oversampling number. Therefore, as long as $(K+r) < L$, the server cannot recover the data of user $i$ (i.e., $\boldsymbol{S}_i$) as $\mw$ is not full row rank. In general, it is for sure that $K<L$ because the objective of dimension reduction is to decrease the data's dimensionality from $L$ to $K$. In the degradation data modeling considered in this paper (and many other applications), $K\ll L$. Given that the oversampling number $r$ is usually very small, the condition $(K+r)<L$ is reliably met.

Computing $(\boldsymbol{S}\boldsymbol{S}^\top)^q{\boldsymbol{S}}\boldsymbol{W}$ in a federated manner is more complex than computing ${\boldsymbol{S}}\boldsymbol{W}$. To address this challenge, we have noticed that $(\boldsymbol{S}\boldsymbol{S}^\top)^q{\boldsymbol{S}}\boldsymbol{W}={\boldsymbol{S}}(\boldsymbol{S}^\top\boldsymbol{S})^q\boldsymbol{W}$. Also, we have observed that given $\boldsymbol{S}=(\boldsymbol{S}_1^\top, \boldsymbol{S}_2^\top,\ldots,\boldsymbol{S}_I^\top)^\top$, $\boldsymbol{S}^\top\boldsymbol{S}=\sum_{i=1}^I\boldsymbol{S}_i^\top\boldsymbol{S}_i$, which allows for a summation across all $I$ users. This insight leads to the following computational steps for $(\boldsymbol{S}\boldsymbol{S}^\top)^q{\boldsymbol{S}}\boldsymbol{W}$: (i) Each user $i$ downloads $\boldsymbol{W}$ from the server, computes $\boldsymbol{S}_i^\top\boldsymbol{S}_i\boldsymbol{W}\in\real^{L\times (K+r)}$ using its local data, and send it to the server; (ii) The server updates $\boldsymbol{W}$ using the formula $\boldsymbol{W}\coloneqq\sum_{i=1}^I\boldsymbol{S}_i^\top\boldsymbol{S}_i\boldsymbol{W}\in\real^{L\times (K+r)}$; (iii) Steps (i) and (ii) are repeated for $q$ times; (iv) The final step, analogous to computing $\boldsymbol{S} \boldsymbol{W}$ discussed earlier, involves each user $i$ downloading the latest $\boldsymbol{W}$ from the server, computing $\boldsymbol{Y}_i \coloneqq \boldsymbol{S}_i \boldsymbol{W}$ locally, and sending $\boldsymbol{Y}_i$ to the server. The server then aggregates these to form $\boldsymbol{Y} = (\boldsymbol{Y}_1^\top, \boldsymbol{Y}_2^\top, ..., \boldsymbol{Y}_I^\top)^\top$. Since, $\boldsymbol{W}$ has been updated for $q$ times, $\boldsymbol{Y}$ represents $(\boldsymbol{S}\boldsymbol{S}^\top)^q \boldsymbol{S} \boldsymbol{W}$. Computing $(\boldsymbol{S}\boldsymbol{S}^\top)^q{\boldsymbol{S}}\boldsymbol{W}$, what each user $i$ sends to the server is $\boldsymbol{S}_i^\top\boldsymbol{S}_i\boldsymbol{W}$. According to Proposition \ref{prop:recovery}, $\boldsymbol{S}_i^\top\boldsymbol{S}_i$ cannot be recovered when the dimensions of $\boldsymbol{W}$ satisfy $(K+r)<L$, which is always true in our degradation modeling (and many other applications). Actually, even if $\boldsymbol{S}_i^\top\boldsymbol{S}_i$ can be recovered, it is generally impossible to recover $\boldsymbol{S}_i$ from it. Thus, the data privacy of each user is protected.  

In the \textit{second} step of RSVD, the goal is to extract a set of orthonormal basis vectors for the identified subspace. This can be efficiently handled by the server, which performs QR decomposition on $\boldsymbol{Y}$. Here, $\boldsymbol{Y}$ represents $\boldsymbol{S}\boldsymbol{W}$ when the rank of $\boldsymbol{S}$ is $K$, and represents $(\boldsymbol{S}\boldsymbol{S}^\top)^q\boldsymbol{S}\boldsymbol{W}$ when the rank of $\boldsymbol{S}$ is greater than $K$.

In the \textit{third} step, the computation of the low-dimensional matrix $\boldsymbol{B} = \boldsymbol{Q}^\top \boldsymbol{S} \in \mathbb{R}^{K \times L}$ is required. Given that $\boldsymbol{S} = (\boldsymbol{S}_1^\top, \boldsymbol{S}_2^\top, \ldots, \boldsymbol{S}_I^\top)^\top$, with each $\boldsymbol{S}_i\in\real^{J_i\times L}$ owned by a different user, direct computation of $\boldsymbol{B}$ is not feasible. To address this issue, we have noticed the following. If we split $\boldsymbol{Q}\in\real^{J\times K}$ into $I$ submatrices according to the number of samples each user has---that is---$\boldsymbol{Q}=(\boldsymbol{Q}_1^\top,\boldsymbol{Q}_2^\top,\ldots,\boldsymbol{Q}_I^\top)^\top$, where $\boldsymbol{Q}_i\in\real^{J_i\times K}$, then $\boldsymbol{B}\coloneqq\sum_{i=1}^I\boldsymbol{Q}_i^\top\boldsymbol{S}_i$. This insight inspires us to propose the following computational steps: (i) The server splits $\boldsymbol{Q}$ into $\{\boldsymbol{Q}_i\}_{i=1}^I$. (ii) each user $i$ downloads $\boldsymbol{Q}_i$ from the server, computes $\boldsymbol{B}_i=\boldsymbol{Q}_i^\top\boldsymbol{S}_i$, and sends $\boldsymbol{B}_i$ to the server. (iii) The server aggregates $\boldsymbol{B}=\sum_{i=1}^I\boldsymbol{B}_i$. Here, what is sent by user $i$ to the server is $\boldsymbol{B}_i\in\real^{(K+r)\times L}$. Based on $\boldsymbol{B}_i^\top=\boldsymbol{S}_i^\top\boldsymbol{Q}_i$ and Proposition \ref{prop:recovery}, as long as $K<J_i$, $\boldsymbol{S}_i^\top$ cannot be recovered by the server. Recall that $K$ represents the number of principal components and $J_i$ is the sample size of user $i$. Thus, $K<J_i$ may not hold since there is no guarantee that a user's sample size is smaller than the number of principal components. To address this challenge, we need a masking server (or one of the users) to generate an orthogonal masking matrix $\boldsymbol{P}\in\real^{K\times K}$ and share it with all users. Each user then masks its $\boldsymbol{B}_i$ with $\boldsymbol{P}$ as follows: $\tilde{\boldsymbol{B}}_i\coloneqq\boldsymbol{P}\boldsymbol{B}_i\in\real^{K\times L}$ and sends it to the server. The server then aggregates $\tilde{\boldsymbol{B}}=\sum_{i=1}^I\tilde{\boldsymbol{B}}_i$ and centers it by ensuring that the average of each column is zero, which provides a centered matrix $\tilde{\boldsymbol{B}}'$. Then, the server conducts compact SVD on $\tilde{\boldsymbol{B}}'$, yielding $\tilde{\boldsymbol{B}}'=\hat{\boldsymbol{U}}\hat{\boldsymbol{\Sigma}}\hat{\boldsymbol{V}}^\top$. Finally, each user $i$ downloads $\hat{\boldsymbol{U}}\in\real^{K\times K}$, $\hat{\boldsymbol{\Sigma}}\in\real^{K\times K}$, and $\hat{\boldsymbol{V}}\in\real^{L\times K}$ from the server. Each user $i$ can compute the left singular vectors, singular values, and right singular vectors of the original uncentered data matrix $\tilde{\boldsymbol{S}}$ as follows: $\boldsymbol{U}=\boldsymbol{Q}\boldsymbol{P}^\top\hat{\boldsymbol{U}}\in\real^{J\times K}$, $\boldsymbol{\Sigma}=\hat{\boldsymbol{\Sigma}}\in\real^{K\times K}$, and $\boldsymbol{V}=\hat{\boldsymbol{V}}\in\real^{L\times K}$. The right singular vectors in $\boldsymbol{V}$ are needed for the computation of local MFPC scores.

The proposed Federated Randomized SVD (FRSVD) algorithm is summarized in Algorithm \ref{algonew:FedRandSVDComplete}. As discussed earlier, the $K$ columns of matrix $\boldsymbol{V}$ are the $K$ eigenvectors (singular vectors) that can be used to compute the MFPC-scores. Recall the degradation signal for asset $j$ of user $i$ is denoted as $\boldsymbol{s}_{ij}\in\mathbb{R}^{L}$, and the MFPC-score ${x}_{ijk}=(\boldsymbol{s}_{ij}-\bar{\boldsymbol{\mu}})^\top\boldsymbol{v}_k$, where $\boldsymbol{v}_k$ is the $k$th column of $\boldsymbol{V}$. However, the signal mean $\bar{\boldsymbol{\mu}}$ is unknown. Therefore, we compute the MFPC-scores using the uncentered signal---that is---${x}_{ijk}=\boldsymbol{s}_{ij}^\top\boldsymbol{v}_k$. It is important to note that using the uncentered signal does not affect the performance of the subsequent regression-based prognostic model. This is because the use of the uncentered signal is equivalent to adding a constant (i.e., $\bar{\boldsymbol{\mu}}^\top\boldsymbol{v}_k$) to the MFPC-scores of assets across all users, which only shifts the intercept of the regression model without influencing its predictive capability.

\begin{algorithm}[h]
  \DontPrintSemicolon
  \SetAlgoLined
\KwInput{Data from $I$ users $\{\ms_i\}_{i=1}^I\in\mathbb{R}^{J_i\times L}$, target number of principal components $K$, oversampling number $r$, and the exponent $q$}
\KwOutput{The left singular vectors $\boldsymbol{U}\in\real^{J\times K}$, singular values $\boldsymbol{\Sigma}\in\real^{K\times K}$, and right singular vectors $\boldsymbol{V}\in\real^{L\times K}$} 

\caption{Federated Randomized SVD (FRSVD)}
\label{algonew:FedRandSVDComplete}
\SetKwFunction{FRName}{FRSVD}
\SetKwProg{Repeat}{repeat}{}{end}
\SetKwProg{IF}{If}{}{end}
\SetKwFunction{Equation}{Equation}
  \SetKwProg{Fn}{Function}{:}{}
  \SetKwProg{Pn}{Function}{:}{end}
  \SetKwProg{Pclient}{Each Client}{:}{end}
\SetKwProg{Pclientone}{Client $1$}{:}{end}
\SetKwProg{Pclientk}{Client $K$}{:}{end}
  \SetKwProg{Pserver}{Server}{:}{end}
    \SetKwProg{Proundone}{Round $1$}{:}{end}
    \SetKwProg{Proundtwo}{Round $2$}{:}{end}
    \SetKwProg{Proundthree}{Round $3$}{:}{end}
  \SetKw{Break}{break}
  \SetKwFunction{SA}{Secure aggregation}

 \begin{enumerate}
     \itemindent=-18pt
     \itemsep=0pt
     \item \textbf{Server} generates a random matrix $\boldsymbol{W}\in\mathbb{R}^{L\times (K+r)}$\\

     \item (i) \textbf{User} $i,i=1,\ldots, I,$ in parallel downloads $\boldsymbol{W}$, computes $\boldsymbol{W}_i\coloneqq\boldsymbol{S}_i^\top\boldsymbol{S}_i\boldsymbol{W}$, and uploads $\boldsymbol{W}_i\in\mathbb{R}^{L\times (K+r)}$ to the server\\
     
     \hspace{-6mm}(ii) \textbf{Server} aggregates $\boldsymbol{W}\coloneqq\sum_{i=1}^I\boldsymbol{W}_i$\\
     
     \hspace{-6mm}(iii) Repeat (i) and (ii) until {$q$} times
  
\item \textbf{User} $i,i=1,\ldots, I,$ in parallel downloads $\boldsymbol{W}$ from the server, computes  $\boldsymbol{Y}_i\coloneqq\boldsymbol{S}_i\boldsymbol{W}$, and uploads $\boldsymbol{Y}_i\in\mathbb{R}^{J_i\times (K+r)}$ to the server\\

\item \textbf{Server} aggregates $\boldsymbol{Y}=(\boldsymbol{Y}_1^\top,\boldsymbol{Y}_2^\top ,\ldots, \boldsymbol{Y}_I^\top)^\top\in\mathbb{R}^{J\times (K+r)}$, where $J=\sum_{i=1}^IJ_i$\\

\item \textbf{Server} constructs a matrix $\boldsymbol{Q}\in\mathbb{R}^{J\times K}$ whose columns form an orthnormal basis for the range of $\boldsymbol{Y}$ via SVD or QR Decomposition\\

\item \textbf{Server} splits $\boldsymbol{Q}\in\mathbb{R}^{J\times K}$ into $\boldsymbol{Q}_1\in\mathbb{R}^{J_1\times K}$, $\boldsymbol{Q}_2\in\mathbb{R}^{J_2\times K}$, $\ldots$, $\boldsymbol{Q}_I\in\mathbb{R}^{J_I\times K}$\\

\item \textbf{Masking Server} generates a masking matrix $\boldsymbol{P}\in\real^{K\times K}$

\item \textbf{User} $i,i=1,\ldots, I,$ in parallel downloads $\boldsymbol{Q}_i$ and the masking matrix $\boldsymbol{P}$, computes  $\tilde{\boldsymbol{B}}_i\coloneqq\boldsymbol{P}\boldsymbol{Q}_i^\top\boldsymbol{S}_i$, and uploads $\tilde{\boldsymbol{B}}_i\in\mathbb{R}^{K\times L}$ to the server\\

\item \textbf{Server} aggregates $\tilde{\boldsymbol{B}}=\sum_{i=1}^I\tilde{\boldsymbol{B}}_i$ and centralizes $\tilde{\boldsymbol{B}}$, yielding $\tilde{\boldsymbol{B}}'$ (the centralization is not required by FRSVD but is necessary for MFPCA). Then, \textbf{Server} performs SVD on $\tilde{\boldsymbol{B}}'$, yielding $\tilde{\boldsymbol{B}}'=\hat{\boldsymbol{U}}\hat{\boldsymbol{\Sigma}}\hat{\boldsymbol{V}}^\top$

\item \textbf{User} $i,i=1,\ldots, I,$ in parallel downloads $\hat{\boldsymbol{U}}\in\real^{K\times K}$, $\hat{\boldsymbol{\Sigma}}\in\real^{K\times K}$, and $\hat{\boldsymbol{V}}\in\real^{L\times K}$ and computes $\boldsymbol{U}=\boldsymbol{P}^\top\boldsymbol{Q}\hat{\boldsymbol{U}}\in\real^{J\times K}$, $\boldsymbol{\Sigma}=\hat{\boldsymbol{\Sigma}}\in\real^{K\times K}$, and $\boldsymbol{V}=\hat{\boldsymbol{V}}\in\real^{L\times K}$
   \end{enumerate} 
\end{algorithm}

\subsection{Communication and Computational Costs}\label{sec:frsvd_cost}

In this subsection, we analyze the communication and computational costs of the proposed FRSVD algorithm and compare them with those of the FSVD algorithm developed in \cite{chai2022practical}. Recall that $\boldsymbol{S}\in\real^{J\times L}$ and in our degradation modeling setting $L\gg J$. If the rank of $\boldsymbol{S}$ is higher than $K$, then the communication cost of the proposed FRSVD algorithm consists of the following steps: (1) Each user downloads $\boldsymbol{W}\in\real^{L\times (K+r)}$ from the server and sends $\boldsymbol{W}_i\in\real^{L\times (K+r)}$ to the server $q$ times, (2) Each user downloads $\boldsymbol{W}\in\real^{L\times (K+r)}$ from the server, (3) Each user uploads $\boldsymbol{Y}_i\in\mathbb{R}^{J_i\times (K+r)}$ to the server, (4) Each user downloads $\boldsymbol{Q}_i\in\mathbb{R}^{J_i\times K}$ from the server, (5) Each user downloads $\boldsymbol{P}\in\real^{K\times K}$ from the masking server, (6) Each user uploads $\tilde{\boldsymbol{B}}_i\in\mathbb{R}^{K\times L}$ to the server, and (7) Each user downloads $\hat{\boldsymbol{U}}\in\real^{K\times K}$, $\hat{\boldsymbol{\Sigma}}\in\real^{K\times K}$ and $\hat{\boldsymbol{V}}\in\real^{L\times K}$ from the server. Since $K+r\ll L$, the total communication cost is determined primarily by steps (1), (2), (6), and (7), which is approximately $((2q+3)K+(2q+1)r)\times L$. For example, taking $L = 1\times 10^5$, $J_i = 5$, $J = 500$, $r = 10$, $K = 90$, and $q = 2$, the total communication cost is around $6.8 \times 10^7$ floating-point numbers. If the rank of $\boldsymbol{S}$ is smaller than or equal to $K$, step (1) is not required. As a result, the total communication cost of the proposed FRSVD algorithm is approximately $(3K + r) \times L$, which equals about $2.8 \times 10^7$ floating-point numbers using the same parameters above. In contrast, the communication cost of the FSVD algorithm developed in \cite{chai2022practical} is approximately $2L\times (L+J)$, which amounts to around $2\times 10^{10}$ floating-point numbers.

    Regarding computational cost, the proposed algorithm is significantly more efficient than the FSVD algorithm in \cite{chai2022practical}. Specifically, our proposed algorithm requires the computation of the SVD on a relatively small matrix, $\boldsymbol{B} \in \mathbb{R}^{K \times L}$, and the SVD (or QR decomposition) on another small matrix, $\boldsymbol{Y} \in \mathbb{R}^{J \times (K + r)}$. In contrast, the FSVD algorithm requires performing SVD on a much larger matrix of size $\mathbb{R}^{J \times L}$. It is known that the computation complexity of SVD on matrix $\boldsymbol{A}\in\mathbb{R}^{m\times n}$ is $\mathcal{O}(\min(m^2n,mn^2))$. As a result, the computation complexity of the proposed FRSVD and FSVD in \cite{chai2022practical} are $\mathcal{O}(8\times 10^8)$ and $\mathcal{O}(2.5\times 10^{10})$, respectively, based on the parameters used above.

\section{Federated (Log)-Location-Scale Regression}\label{sec:LLS}
 
In this section, we propose federated (log)-location-scale (LLS) regression that enables multiple users to jointly construct a prognostic model for failure time prediction. After applying the federated dimension reduction method proposed in Section \ref{sec:fdr}, the high-dimensional degradation signals of the $j$th asset of user $i$ (i.e., $\boldsymbol{s}_{ij}\in\mathbb{R}^{M}$) are transformed into low-dimensional features (i.e., MFPC-scores $\{{x}_{ijk}\}_{k=1}^K$). Let\\ $\boldsymbol{x}_{ij}=(1, {x}_{ij1},{x}_{ij2},\ldots,{x}_{ijK})^\top\in\mathbb{R}^{(K+1)}$ represent the vector of features for the $j$th asset of user $i$ and denote the corresponding TTF as $y_{ij}\in\mathbb{R}$, the location-scale regression model can be expressed as follows:

\begin{equation}\label{eq:ls}
y_{ij} = \boldsymbol{x}_{ij}^\top\boldsymbol{\beta} +\sigma\epsilon_{ij},
\end{equation}

\noindent where $\boldsymbol{\beta}=(\beta_0, \beta_1, \beta_2,...,\beta_K)^\top \in \mathbb{R}^{(K+1)}$ is the coefficient vector. $\sigma$ is the scale parameter, and $\epsilon_{ij}$ is
the random noise term with a standard location-scale density
$f(\epsilon)$. For example $f(\epsilon)=1/\sqrt{2\pi}\exp(-\epsilon^2)$ for normal distribution and $f(\epsilon)=\exp(\epsilon-\exp(\epsilon))$ for SEV distribution. As a result, $y_{ij}$ has a density in the form of $\frac{1}{\sigma}f(\frac{y_{ij} - \boldsymbol{x}_{ij}^\top\boldsymbol{\beta}}{\sigma})$. It is well established that if $y_{ij}$ follows a log-location-scale distribution (e.g., log-normal, log-logistic, Weibull), then $\ln(y_{ij})$ follows a location-scale distribution (e.g., normal, logistic, SEV). Therefore, the regression model described in Equation \eqref{eq:ls} can be applied to model TTF data originating from a log-location-scale distribution as well. As a result, we refer to this general modeling approach as (log)-location-scale (LLS) regression.

Given the MFPC-scores and TTFs from all users (i.e., $\{\{\boldsymbol{x}_{ij}\}_{j=1}^{J_i}\}_{i=1}^I$ and $\{\{y_{ij}\}_{j=1}^{J_i}\}_{i=1}^I$), the parameters in LLS regression can be estimated using Maximum Likelihood Estimation (MLE) by solving $\arg\min_{\boldsymbol{\beta}, \sigma}\ell(\boldsymbol{\beta}, \sigma)$, where 

\vspace{-3mm}
\begin{equation}\label{eq:mle}
\ell(\boldsymbol{\beta}, \sigma)=-\sum_{i=1}^I\sum_{j=1}^{J_i}\left(\frac{1}{\sigma}+f\left(\frac{y_{ij} -\boldsymbol{x}_{ij}^\top\boldsymbol{\beta}}{\sigma}\right)\right).
\end{equation}

Usually, the following reparameterization is carried out to change the optimization objective function to be convex: $\tilde{\sigma}=1/\sigma$, $\tilde{\boldsymbol{\beta}}=\boldsymbol{\beta}/\sigma$, which yields solving $\min_{\tilde{\boldsymbol{\beta}}, \tilde{\sigma}}\tilde{\ell}(\tilde{\boldsymbol{\beta}}, \tilde{\sigma})$, where $\tilde{\ell}(\tilde{\boldsymbol{\beta}}, \tilde{\sigma})=-\sum_{i=1}^I\sum_{j=1}^{J_i}(\tilde{\sigma}+f(\tilde{\sigma}y_{ij} -\boldsymbol{x}_{ij}^\top\tilde{\boldsymbol{\beta}}))$. For notation simplicity, it is common to define $\boldsymbol{\theta}\coloneqq(\tilde{\sigma},\tilde{\boldsymbol{\beta}}^\top)^\top$, and thus the optimization criterion can be expressed as

\vspace{-3mm}
\begin{equation}\label{eq:mle2}
\arg\min_{\boldsymbol{\theta}}\tilde{\ell}(\boldsymbol{\theta}),
\end{equation}

\noindent where $\tilde{\ell}(\boldsymbol{\theta})=-\sum_{i=1}^I\sum_{j=1}^{J_i}(\tilde{\sigma}+f(\tilde{\sigma}y_{ij} -\boldsymbol{x}_{ij}^\top\tilde{\boldsymbol{\beta}}))$. 

Criterion \eqref{eq:mle2} can be solved using convex optimization software packages if the MFPC-scores and TTFs of the assets from all users could be aggregated. However, this is not feasible due to privacy constraints, as the MFPC-scores and TTFs of each user cannot be shared with other users or a central server. To address this challenge, we propose a federated parameter estimation algorithm based on Gradient Descent (GD). 

GD is an optimization algorithm used to minimize a cost function by iteratively adjusting the parameters of a model in the direction of the negative gradient. To optimize criterion \eqref{eq:mle2}, GD iteratively updates the unknown parameters by using $\boldsymbol{\theta}^{(l)}=\boldsymbol{\theta}^{(l-1)}-\alpha\nabla\tilde{\ell}(\boldsymbol{\theta}^{(l-1)})$ until convergence, where $l$ is the step index, $\alpha$ is the learning rate that controls the size of the steps, and $\nabla\tilde{\ell}(\boldsymbol{\theta}^{(l-1)})=(\frac{\partial \tilde{\ell}(\boldsymbol{\theta}^{(l-1)})}{\partial\tilde{\sigma}},\frac{\partial\tilde{\ell}(\boldsymbol{\theta}^{(l-1)})}{\partial\tilde{\beta}_0},\ldots,\frac{\partial\tilde{\ell}(\boldsymbol{\theta}^{(l-1)})}{\partial\tilde{\beta}_K})^\top$ is the partial gradient at $\boldsymbol{\theta}^{(l-1)}$.

The $\tilde{\ell}(\boldsymbol{\theta})$ in criterion \eqref{eq:mle2} can be rewritten as $\tilde{\ell}(\boldsymbol{\theta})=\sum_{i=1}^I\tilde{\ell}_i(\boldsymbol{\theta})$, where\\ $\tilde{\ell}_i(\boldsymbol{\theta})=-\sum_{j=1}^{J_i}\left(\tilde{\sigma}+f\left(\tilde{\sigma}y_{ij} -\boldsymbol{x}_{ij}^\top\tilde{\boldsymbol{\beta}}\right)\right)$ is the objective function constructed using the data of user $i$. Similarly, the gradient $\nabla\tilde{\ell}(\boldsymbol{\theta}^{(l-1)})=\sum_{i=1}^I\nabla\tilde{\ell}_i(\boldsymbol{\theta}^{(l-1)})$, where $\nabla\tilde{\ell}_i(\boldsymbol{\theta}^{(l-1)})=(\frac{\partial \tilde{\ell}_i(\boldsymbol{\theta}^{(l-1)})}{\partial\tilde{\sigma}},\frac{\partial\tilde{\ell}_i(\boldsymbol{\theta}^{(l-1)})}{\partial\tilde{\beta}_0},\ldots,\frac{\partial\tilde{\ell}_i(\boldsymbol{\theta}^{(l-1)})}{\partial\tilde{\beta}_K})^\top$ is the partial gradient computed by user $i$. As a result, the GD updating equation can be expressed as $\boldsymbol{\theta}^{(l)}=\boldsymbol{\theta}^{(l-1)}-\alpha\sum_{i=1}^I\nabla\tilde{\ell}_i(\boldsymbol{\theta}^{(l-1)})$. This implies that when using GD for parameter estimation, the total gradient is the sum of the gradients computed by each user. This allows us to develop a federated parameter estimation algorithm, where each user computes its local gradient using its own data and shares the local gradient with a central server. The server then aggregates the gradients from all users and performs the parameter updates. Specifically, the proposed federated algorithm works as follows: In the \textit{first} step, all the parameters are initialized by drawing from a uniform distribution over the interval (0, 1), i.e., $ \boldsymbol{\theta}^{(0)} \sim \mathcal{U}(0,1)^{K+2}$. Then, in the \textit{second} step, (i) The server randomly initializes the parameter vector $\boldsymbol{\theta}^{(l-1)}=(\tsigma^{(l-1)},\tbeta^{(l-1)\top})^\top$, where the iteration index $l=1$; (ii) Each user downloads the parameters $\boldsymbol{\theta}^{(l-1)}$ from the server and computes their own local gradients $\nabla\tilde{\ell}_i(\boldsymbol{\theta}^{(l-1)})$. Proposition \ref{proposition2} provides the gradients for the parameters of three commonly used location-scale distributions. (iii) Each user sends its local gradients $\nabla\tilde{\ell}_i(\boldsymbol{\theta}^{(l-1)})$ to the central server, which aggregates the gradients from all users $\nabla\tilde{\ell}(\boldsymbol{\theta}^{(l-1)})=\sum_{i=1}^I\nabla\tilde{\ell}_i(\boldsymbol{\theta}^{(l-1)})$ and uses the aggregated gradients to update the parameters $\boldsymbol{\theta}^{(l)}=\boldsymbol{\theta}^{(l-1)}-\alpha\nabla\tilde{\ell}(\boldsymbol{\theta}^{(l-1)})$, and then increments $l\leftarrow l+1$. Steps (ii) and (iii) are repeated until the number of iterations reaches the maximum $M^*$, or the change in parameters between two iterations $\Delta \boldsymbol{\theta}\coloneqq\|\boldsymbol{\theta}^{(l)}-\boldsymbol{\theta}^{(l-1)}\|_2$ falls below a convergence tolerance $\delta$. In the \textit{third} step, each user downloads the optimal parameters $\boldsymbol{\theta}^{(l)}=(\tilde{\sigma}^{(l)},\tilde{\boldsymbol{\beta}}^{(l)\top})^\top$, and we denote it as $\boldsymbol{\theta}^*=(\tilde{\sigma}^*,\tilde{\boldsymbol{\beta}}^{*\top})^\top$.

\begin{Proposition}\label{proposition2}
Recall that the low-dimensional features and the corresponding TTF for the $j$th asset of user $i$ are denoted as $\boldsymbol{x}_{ij}$ and $y_{ij}$, respectively. For simplicity, we use $\boldsymbol{\theta}$ in place of $\boldsymbol{\theta}^{(l)}$. The gradient of the negative log-likelihood for user $i$ with respect to $\tbeta$ and $\tsigma$ is given by:

(i) Normal Distribution: 
\begin{flalign*}
&\begin{aligned}
\begin{cases}\frac{\partial \lp}{\partial \tsigma}
=\frac{-J_i}{\tsigma}+\sum_{j=1}^{J_i}(\tsigma y_{ij}-\tcombn)y_{ij}\\
\frac{\partial \lp}{\partial \tbeta}=-\sum_{j=1}^{J_i}(\tsigma y_{ij}-\tcombn)\boldsymbol{x}_{ij}^\top\end{cases}
\end{aligned}&&
\end{flalign*}

(ii) SEV Distribution:
\begin{flalign*}
&\begin{aligned}
\begin{cases}
\frac{\partial \lp}{\partial \tsigma}
=\frac{-J_i}{\tsigma}-\sum_{j=1}^{J_i}\Big\{(1-e^{y_{ij}\tsigma-\tcombn})y_{ij}\Big\}\\
\frac{\partial \lp}{\partial \tbeta}=\sum_{j=1}^{J_i}\Big\{\Big[1-e^{y_{ij}\tsigma-\tcombn}\Big] \boldsymbol{x}_{ij}^\top\Big\}\end{cases}
\end{aligned}&&
\end{flalign*}

(iii) Logistic Distribution:
\begin{flalign*}
&\begin{aligned}
\begin{cases}\frac{\partial \lp}{\partial \tsigma}
=\sum_{j=1}^{J_i}\Bigg\{y_{ij}\Bigg[\frac{2e^{y_{ij}\tsigma-\tcombn}}{1+e^{y_{ij}\tsigma-\tcombn}}-1 \Bigg]  \Bigg\}-\frac{J_i}{\tsigma}\\ 
\frac{\partial \lp}{\partial \tbeta}=\sum_{j=1}^{J_i}\Bigg\{\Bigg[1-2\frac{e^{y_{ij}\tsigma-\tcombn}}{1+e^{y_{ij}\tsigma-\tcombn}} \Bigg] \boldsymbol{x}_{ij}^\top \Bigg\} \end{cases}
\end{aligned}&&
\end{flalign*}
\end{Proposition}

\noindent The derivation details can be found in Section \ref{proof2}.

We summarize the proposed federated (log)-location-scale (LLS) regression algorithm in Algorithm \ref{algo:FedGDLLS}.

\begin{algorithm}[h]

  \DontPrintSemicolon
  \SetAlgoLined
\KwInput{MFPC-scores $\{\{\boldsymbol{x}_{ij}\}_{j=1}^{J_i}\}_{i=1}^I\in \real^{K+1}$ and the corresponding TTF $\{\{y_{ij}\}_{j=1}^{J_i}\}_{i=1}^I\in \real$, convergence tolerance $\delta$, learning rate $\alpha$, and maximum iteration number $M^*$
}
\KwOutput{Optimal parameter $\boldsymbol{\theta}^*=(\tilde{\sigma}^*,\tilde{\boldsymbol{\beta}}^{*\top})^\top$} 

\vspace{2mm}
\caption{Federated LLS Regression}
\label{algo:FedGDLLS}
 \begin{enumerate}
     \itemindent=-18pt
     \itemsep=0pt
     \item \textbf{Server} initializes parameters $\boldsymbol{\theta}^{(0)}=(\tilde{\sigma}^{(0)},\tilde{\boldsymbol{\beta}}^{(0)\top})^\top$, where $\tilde{\sigma}^{(0)}\sim\mathcal{U}(0,1)$, and $\tilde{\boldsymbol{\beta}}^{(0)}\sim\mathcal{U}(0,1)^{K+1}$ \\

     \item Let iteration index $l=1$\\
     \hspace{-6mm}(i)  \textbf{User} $i,i=1,\ldots, I,$ in parallel downloads $\boldsymbol{\theta}^{(l-1)}$ from the server, computes local gradients $\nabla\tilde{\ell}_i(\boldsymbol{\theta}^{(l-1)})$, and sends it to the server
     \\
     
     \hspace{-6mm}(ii) \textbf{Server} aggregates $\nabla\tilde{\ell}(\boldsymbol{\theta}^{(l-1)})=\sum_{i=1}^I\nabla\tilde{\ell}_i(\boldsymbol{\theta}^{(l-1)})$ and updates $\boldsymbol{\theta}^{(l)}=\boldsymbol{\theta}^{(l-1)}-\alpha\nabla\tilde{\ell}(\boldsymbol{\theta}^{(l-1)})$. Then, \textbf{Server} computes $\Delta \boldsymbol{\theta}=\|\boldsymbol{\theta}^{(l)}-\boldsymbol{\theta}^{(l-1)}\|_2$. If $\Delta\boldsymbol{\theta}<\delta$, break the iteration; otherwise, let $l\leftarrow l+1$.\\
     
     \hspace{-6mm}(iii) Repeat (i) and (ii) until the iteration number reaches $M^*$
  
\item \textbf{User} $i,i=1,\ldots, I,$ in parallel downloads $\boldsymbol{\theta}^{(l)}=(\tilde{\sigma}^{(l)},\tilde{\boldsymbol{\beta}}^{(l)\top})^\top$\\
\end{enumerate}

\end{algorithm}

\section{Simulation Study}\label{section3.5}
In this section, we conduct simulation experiments to evaluate the effectiveness of the proposed two-stage federated prognostics model. The experiments were performed on an Apple M3 Max processor with a 14-core CPU and 36GB of RAM.

\subsection{Data Generation Settings and Benchmarks}\label{section.simu.1}

Recall that we have denoted the number of users by $I$ and the number of training samples for the $i$th user by $J_i$, $i=1,\ldots, I$. In this study, we set $I=100$ and let $J_i$ follow a discrete uniform distribution $\mathcal{U}(2,20)$. In other words, the number of training samples per user varies between 2 and 20. The degradation signals and TTFs for each user are generated using the same procedure. Taking user $i$ as an example, we assume that the underlying degradation paths follow the form $s_{ij}(t)=-c_{ij}/\ln(t)$, where $c_{ij}\sim \mathcal{N}(1, 0.25^2)$, $0\leq t<1$, and $j=1, 2,\cdots, J_i$. The TTF is determined by the time that the underlying degradation path $s_{ij}(t)$ reaches a threshold $D=2$ plus some noise. In other words, the logarithmic failure time $\ln(y_{ij}) = -c_i/D+\varepsilon_{ij}$, where $\varepsilon_{ij}\sim \mathcal{N}(0,0.025^2)$. As a result, the TTF follows a lognormal distribution. It is known that the observed degradation signals are discrete, noisy measurements of the underlying degradation path. Therefore, we generate the observed degradation signals for the $j$th sample of user $i$ as follows: $s_{ij}(\tau_{ij}) = -c_{ij}/\ln(\tau_{ij})+\epsilon_{ij}(\tau_{ij})$, where $\tau_{ij} = 0.001, 0.002, \ldots, \floor*{y_{ij}/0.001}\times0.001$ ($\floor*{\cdot}$ is the floor operator) and $\epsilon_{ij}(\tau_{ij})\sim \mathcal{N}(0,0.05^2)$. To increase the complexity of the degradation data, each degradation signal is randomly truncated before its failure time. The truncation time is determined by $\ceil*{z\times \floor*{y_{ij}/0.001}}$, where $z\sim Beta(2,3)$, and $\ceil*{\cdot}$ denotes the ceiling operator. 

We generate $50$ test samples following the same procedure described above, except for the truncation time. The truncation times of the test samples are evenly assigned to $10\%, 20\%, \cdots, 90\%, 95\%$ of their TTFs. In other words, $5$ samples are truncated at $10\%$ of their TTFs, $5$ samples are truncated at $20\%$, etc.

We refer to our proposed federated prognostic model as \textit{Federated with RSVD} and compare its performance to three benchmarks: (i) \textit{Non-Federated with RSVD}, (ii) \textit{Non-Federated with SVD}, and (iii) \textit{Individual}. For the first benchmark, we begin by aggregating the training samples from all 100 users. Next, we perform MFPCA on the aggregated signals using RSVD to extract features. These extracted features are then regressed against TTFs to build a lognormal regression model, with parameters estimated using standard gradient descent. It is important to note that the first benchmark is nearly identical to our proposed model, with the key difference being that our method employs federated RSVD and federated LLS regression, whereas the first benchmark uses RSVD and classic LLS regression. The second benchmark is the same as the first benchmark except that it employs classic SVD instead of RSVD. The third benchmark allows each user to train a local prognostic model using only their own training data. Similar to the second benchmark, it utilizes classic SVD and LLS regression.

For the proposed method and the first benchmark, we set $q=2, r=10$ for RSVD. In addition, we apply the adaptive fraction-of-variance-explained (FVE) method proposed by \cite{fang2017scalable} to determine the number of MFPC-scores from RSVD, with the threshold set at $95\%$. For benchmarks (ii) and (iii), the number of MFPC-scores is also determined using the FVE method, with the same $95\%$ threshold. The performance of the proposed method and all three benchmarks is evaluated using the same 50 test samples generated earlier, and the prediction errors are calculated as $|\hat{y}_i^{(t)}-y_i^{(t)}|/|y_i^{(t)}|$, where $\hat{y}_i^{(t)}$ and $y_i^{(t)}$ are the predicted and the true TTF of the $i$th sample in the test data, respectively. 

Since test signals and training signals may have different lengths due to truncation, but MFPCA requires all signals to be of the same length, we employ the adaptive framework proposed by \cite{fang2015adaptive}. Specifically, all training signals are truncated based on the length of the test signal. If the length of a training signal before truncation is shorter than that of the test signal, it is excluded from the training process. This ensures that the test signal and the remaining training signals have the same length. Please note that in some cases, there may be only one or even zero signals left for model training for certain users. When this occurs, the third benchmark model cannot be established, which means that the user cannot train a prognostic model using only its own data. In this case, if only one training sample remains, we set the predicted TTF as either the TTF of the training sample or the current length of the test signal, whichever is larger. If no training samples remain, the predicted TTF is set as the current length of the test signal.

\subsection{Prediction Performance}\label{section.simu.2}
\begin{figure}[ht]
	\centering
	\includegraphics[scale=0.6]{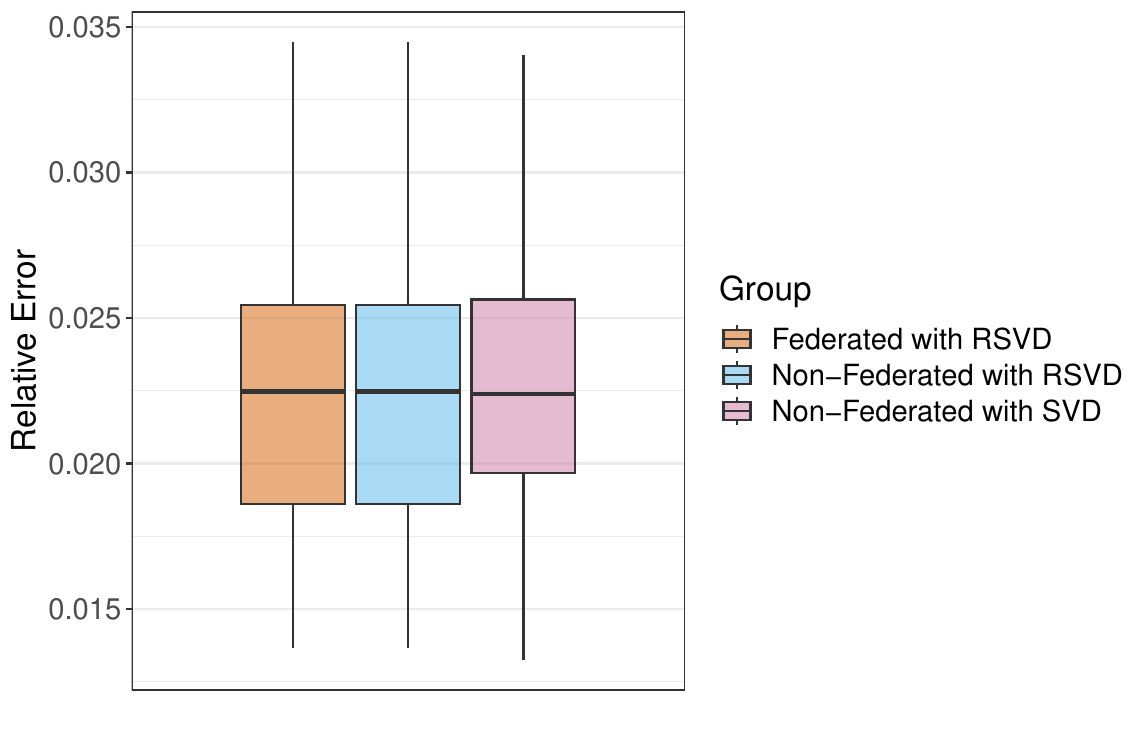}
  \caption{Prediction errors of the proposed method and benchmarks (i) and (ii) (left: \textit{Federated with RSVD (Proposed)}; middle: \textit{Non-Federated with RSVD}; right: \textit{Non-Federated with SVD}).}
	\label{figure.simu.1}
\end{figure}

We first present the prediction errors of the proposed method (i.e., \textit{Federated with RSVD}) and the benchmarks (i) (i.e., \textit{Non-Federated with RSVD}) and (ii) (i.e., \textit{Non-Federated with SVD}) in Figure \ref{figure.simu.1}. The results in Figure \ref{figure.simu.1} show that the proposed method (\textit{Federated with RSVD}) and benchmark (i) (\textit{Non-Federated with RSVD}) achieve identical performance in terms of prediction accuracy. The proposed method utilizes the federated RSVD algorithm for feature extraction and the federated LLS regression for prognostic model construction, while benchmark (i) employs classic RSVD and LLS regression. Therefore, we conclude that the federated RSVD and federated LLS regression do not compromise the performance compared to the traditional RSVD and LLS regression methods. This result is expected because the federated RSVD is a redesign of the classic RSVD algorithm that enables multiple users to collaboratively perform RSVD while maintaining data privacy and confidentiality without compromising RSVD accuracy. Similarly, the federated LLS regression matches the performance of the classic LLS regression, as it leverages the fact that the overall gradient is the sum of gradients from all users. 

Figure \ref{figure.simu.1} also indicates that the performance of the proposed method (and benchmark (i)) is comparable to that of benchmark (ii) (\textit{Non-Federated with SVD}), with benchmark (ii) performing slightly better. The median (and IQR) of the prediction errors for the proposed method and benchmark (i) is 0.0225 (0.007), compared to 0.0224 (0.006) for benchmark (ii). Recall that the proposed method and benchmark (i) use RSVD, while benchmark (ii) employs the classic SVD. This slight difference is expected, as RSVD is an approximation of SVD, albeit with minimal approximation errors.

\begin{figure*}[ht]
	\centering
	\includegraphics[width=6.1in]{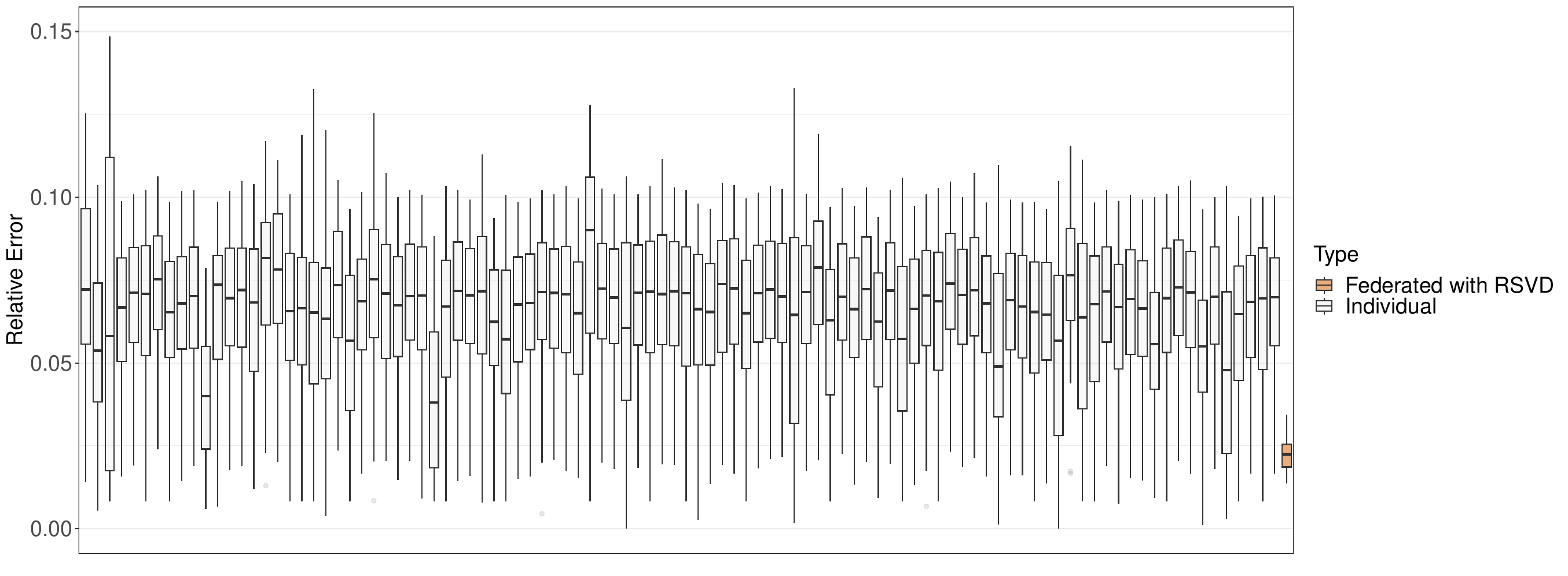}
  \caption{Comparison of prediction errors for 100 local models trained by each of the 100 \textit{individual} users and the proposed federated prognostic model.} 
	\label{figure.simu.3}
\end{figure*}

Figure \ref{figure.simu.3} compares the prediction accuracy of the proposed federated prognostic model with that of benchmark (iii), \textit{Individual}, where each user trains a local prognostic model using only its own training data. The results show that the proposed federated prognostic model significantly outperforms the individual models trained by each user. For instance, the median (and IQR) of the prediction errors for the proposed method is 0.0225 (0.007), whereas the median (and IQR) for the best-performing individual user is 0.0381 (0.041). This is expected, as the proposed method leverages training data from all users, while benchmark (iii) relies solely on each user’s own data. These results highlight the advantages of participating in the construction of the federated prognostic model.

\subsection{Communication and Computation Costs}
\label{sec:simulation.cost}

In this subsection, we examine the computational time and communication cost of the proposed \textit{Federated with RSVD (FRSVD)} method and compare it with the \textit{Federated with SVD (FSVD)} method proposed in \cite{chai2022practical}. We consider different numbers of users, ranging from 100 to 800. As discussed in Section \ref{section.simu.1}, the number of training samples per user is drawn from a uniform distribution and varies between 2 and 20. We report the computation times of \textit{FRSVD} and \textit{FSVD} in Table \ref{table.computation}. 

Table \ref{table.computation} shows that the computational time of the proposed \textit{FRSVD} is significantly smaller than that of \textit{FSVD}. The computational times for \textit{FRSVD} (and \textit{FSVD}) are $0.638 (2.034)$, $1.615 (10.928)$, $2.559 (30.239)$, $3.286 (62.111)$, $3.969 (109.351)$, $ 4.992 (182.702)$, $5.876 (277.239)$, and $6.669 (405.790)$ when the number of users is $100, 200, 300, 400, 500, 600, 700$, and $800$, respectively. These findings align with our analysis in Section \ref{sec:frsvd_cost} , as the proposed \textit{FRSVD} performs SVD on matrices that are significantly smaller compared to those used in \textit{FSVD}.
It is important to note that in this study, the communication cost (i.e., data transmission time between users and the server) is not considered, as the entire process was simulated on a single device and no data on actual transmission times across multiple devices is available. However, the communication cost of the proposed \textit{FRSVD} is expected to be significantly lower than that of \textit{FSVD}, as the sizes of the matrices transmitted between users and the server in \textit{FRSVD} are much smaller compared to those transmitted in \textit{FSVD} (see Section \ref{sec:frsvd_cost} for details).

\begin{table}[htbp]
\centering
\caption{Comparison of computational time between the model using the \textit{Federated with RSVD (FRSVD)} algorithm and the model using the \textit{Federated with SVD (RSVD)} algorithm across different numbers of local users}
\begin{tabular}{rrrr}
\hline
&\multicolumn{2}{c}{\text{ Time (min)}} \\
\cline{2-3}Number of Users
&  \textit{FRSVD}& \textit{FSVD} \\ \hline 
100& 0.638 & 2.034\\
  200&  1.615&10.928\\
300& 2.559 &30.239 \\
400& 3.286 &62.111\\
 500&3.969&109.351\\
  600&4.992& 182.702 \\
700&5.876&277.239 \\
800& 6.669& 405.790\\ 
\hline
\label{table.computation}
\end{tabular}
\end{table}

\section{Case Study}\label{section4}
In this section, we evaluate the performance of the proposed federated prognostic model, along with several benchmarks, using a dataset from the NASA prognostics data repository.

\subsection{Data Description and Experimental Settings}
The dataset used in this study is the Commercial Modular Aero-Propulsion System Simulation (C-MAPSS) FD001 dataset \cite{saxena2008damage}, a widely recognized dataset for system prognostics and health management. The dataset contains multi-stream time-series signals that capture the degradation of aircraft gas turbine engines. It includes: (i) degradation signals from 100 training engines along with their failure times, (ii) degradation signals from 100 test engines that have been randomly truncated before failure, and (iii) the actual failure times of the 100 test engines. Each engine is monitored by 21 sensors. Since 7 of these sensors provide flat (non-informative) degradation signals, they are removed from the dataset, leaving 14 sensors (sensors 2, 3, 4, 7, 8, 9, 11, 12, 13, 14, 15, 17, 20, and 21).

In this case study, we consider a scenario involving three distinct users (i.e., $M=3$). We randomly allocate the 100 training engines into three different groups. These groups vary significantly in size, with the first group containing 10 engines, the second group 30 engines, and the largest group encompassing 60 engines. This allocation assigns each group as the local data repository for users 1, 2, and 3, respectively. Following the suggestion of \cite{fang2017multistream}, we choose to implement the LLS regression with log-normal distribution. As in the simulation study described in Section \ref{section3.5}, we also utilize the adaptive framework proposed by \cite{fang2015adaptive} to ensure that the test and training signals are of the same length. Specifically, we select only those training signals that are longer than a given test signal. These selected training signals are then truncated to match the length of the test signal. Furthermore, considering the multiple sensors in this dataset, the truncated degradation signals from all sensors are concatenated into a single extended signal \cite{fang2017multistream}. Dimension reduction is subsequently applied to this extended signal using either the proposed federated dimension reduction method or regular MFPCA, as used in the benchmarks. The criteria for selecting the number of principal components to retain and the hyperparameters for RSVD are consistent with those outlined in Section \ref{section3.5}. In addition, we employ the same three benchmarking methods used in Section \ref{section3.5}, namely, (i) \textit{Non-Federated with RSVD}, (ii) \textit{Non-Federated with SVD}, and (iii) \textit{Individual}. 

\subsection{Results and Analysis}

\begin{figure}[ht]
	\centering
	\includegraphics[scale=0.6]{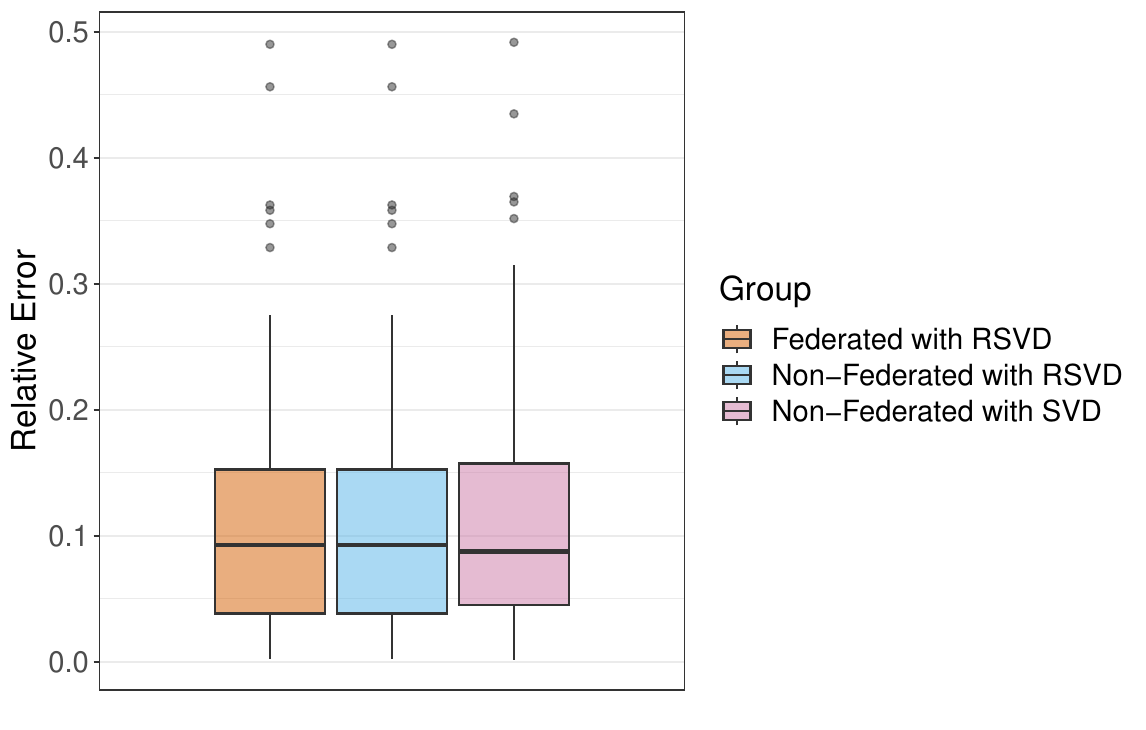}
  \caption{Prediction errors of the proposed method and benchmarks (i) and (ii) (left: \textit{Federated with RSVD (Proposed)}; middle: \textit{Non-Federated with RSVD}; right: \textit{Non-Federated with SVD}).}
	\label{fig:casestudy:bech12}
\end{figure}

\begin{figure}[ht]
	\centering
	\includegraphics[scale=0.6]{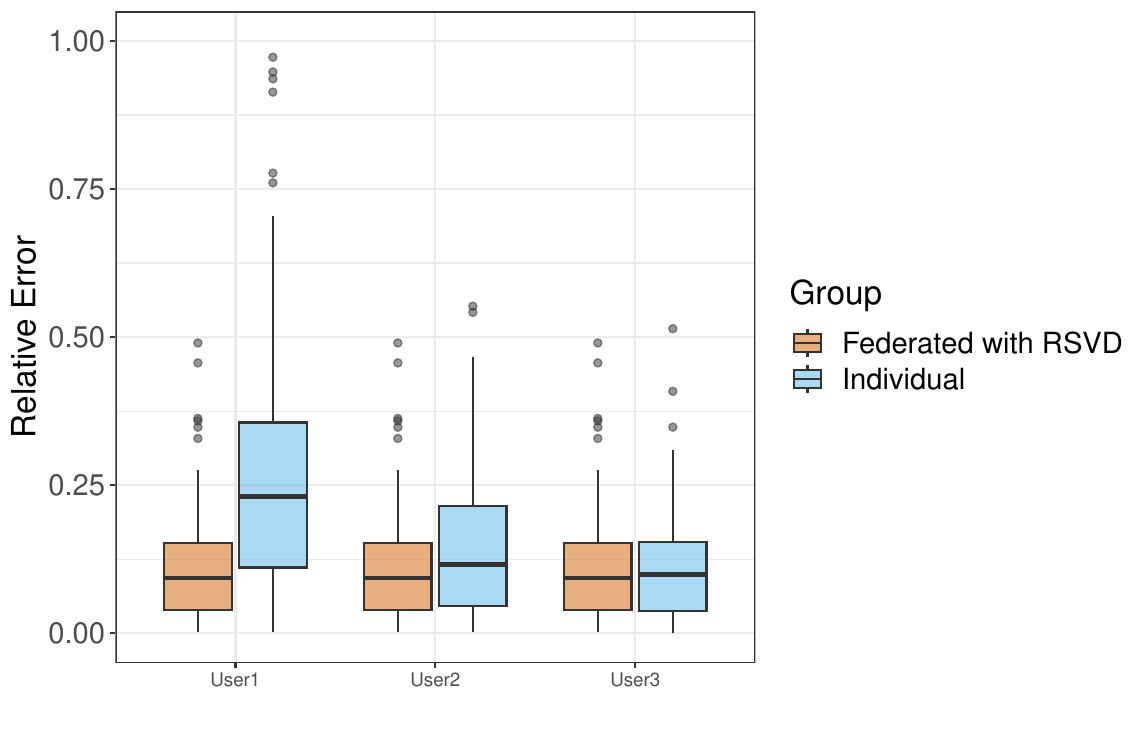}
  \caption{Comparison of relative error by group for each local user of the real dataset (left: \textit{Federated with RSVD (Proposed)}; right: \textit{Individual}).}
	\label{fig:casestudy:bech3}
\end{figure}

We summarize the prediction errors of the proposed federated prognostic model and benchmarks (i) and (ii) in Figure \ref{fig:casestudy:bech12}. The figure shows that the proposed method performs identically to benchmark (i), \textit{Non-Federated with RSVD}, with both methods having a median (IQR) of $0.0928 (0.114)$. The key difference between the two methods is that the proposed model uses federated RSVD and federated LLS regression, while benchmark (i) uses traditional RSVD and LLS regression. Despite this difference, the results indicate that the federated algorithms achieve the same performance as the traditional algorithms, suggesting that the proposed federated approach can match the accuracy of its non-federated counterpart while providing the added benefits of collaborative learning and preserving data privacy. Figure \ref{fig:casestudy:bech3} also shows that benchmark (i) (and the proposed federated model) perform comparably to benchmark (ii). The median (IQR) for benchmark (i) and the proposed method is $0.0928(0.114)$, while benchmark (ii), which uses regular SVD instead of RSVD, has a median (IQR) of $0.0876(0.112)$. This suggests that RSVD, as an approximation of SVD, achieves nearly the same performance as SVD.

We also assess the performance of the proposed federated prognostic model in comparison with benchmark (iii) (i.e., \textit{Individual}) as shown in Figure \ref{fig:casestudy:bech3}. In benchmark (iii), each user trains a local model using their own data and applies this model to predict failure times. It is important to note that both the individual models in benchmark (iii) and the proposed prognostic model utilize the same test dataset. 

The results presented in Figure \ref{fig:casestudy:bech3} reveal that the median (IQR) prediction errors for the individual models trained by Users 1, 2, and 3 are $0.2383(0.304), 0.1155(0.169)$, and $0.0992(0.117)$ respectively, while the median (IQR) for the proposed method is $0.0928(0.114)$. This analysis indicates that the performance of the proposed method significantly surpasses that of the model trained by User 1, exceeds that of the model trained by User 2, and is comparable to the model trained by User 3.

Recalling the local training sample sizes of Users 1, 2, and 3, which are 10, 30, and 60 respectively, these findings suggest that users with limited training samples gain more benefits from participating in the construction of the proposed federated prognostic model. Conversely, users with larger training samples see less incremental benefit from the federated approach. This observation aligns with principles of statistical learning, which suggest that increasing the sample size typically enhances model performance significantly when the initial sample size is small. However, the benefits tend to diminish when the sample size is already sufficiently large.

\section{Conclusions}\label{section5}
In this paper, we proposed a statistical learning-based federated model for industrial prognostics, tackling key challenges such as limited data availability and privacy concerns that hinder collaborative model training across organizations. Our approach comprises two main stages: federated dimension reduction and federated log-location-scale regression. In the first stage, we employed multivariate functional principal component analysis to fuse multi-stream high-dimensional degradation signals and extract low-dimensional MFPC-scores. To enable multiple users to jointly compute MFPC-scores while preserving data confidentiality and locality, we developed a novel federated randomized singular value decomposition (RSVD) algorithm. The proposed federated RSVD algorithm is computationally efficient and has low communication overhead, making it suitable for large-scale local datasets. In the second stage, we used log-location-scale regression to construct a prognostic model mapping an asset's time-to-failure to its MFPC-score and proposed a federated algorithm based on gradient descent for parameter estimation. The proposed federated LLS regression allows multiple organizations to collaboratively construct a prognostic model for the prediction of failure time distributions without sharing raw data.

The proposed model demonstrated significant advantages over existing federated prognostic methods by using statistical learning techniques that perform well with smaller datasets and provide comprehensive failure time distributions. Furthermore, the federated RSVD algorithm greatly reduces communication and computational costs compared to traditional SVD approaches, making the method more scalable and efficient for large-scale industrial applications. Validation using simulated data and a dataset from the NASA repository confirmed the effectiveness, precision, and efficiency of the model, highlighting its potential for practical implementation in equipment health management. 

\section{Appendix} \label{appendix}
\subsection{Proof of Proposition 1}\label{proof1}
Assume $\mz\in\real^{n\times k}, \ms\in\real^{n\times m}, \mg\in\real^{m\times k}$, then from $\mz=\ms\mg$ we have $\mz^\top=\mg^\top\ms^\top$, where $\mg^\top\in\real^{k\times m}.$ 
We will prove the proposition by contradiction. First, we show that if $\ms^\top$ can be uniquely recovered, then the null space of $\mg^\top$ must be  $\{\mathbf{0}\}$ . We then demonstrate that $\Null(\mg^\top) = \{\mathbf{0}\}$ is equivalent to $\mg^\top$ having full column rank.

Suppose the proposition does not hold, meaning that we can uniquely recover $\ms^\top$. This implies that $\ms^\top$ is known and is the unique solution to $\mz^\top = \mg^\top \ms^\top$, given that both $\mz^\top$ and $\mg^\top$ are known.
Let $\vv_1, \vv_2, \cdots, \vv_k$ be $k$ arbitrary vectors from $\Null(\mg^\top)$, and $\mv=(\vv_1, \vv_2, \cdots, \vv_k)$, then $\mg^\top\mv=\mathbf{O}_{k\times k}$. 
Suppose $\ms^*$ is the true data matrix to be recovered, we have $\mg^\top\ms^{*\top}=\mz^\top$. From $\mg^\top(\ms^{*\top}+\mv)=\mg^\top\ms^{*\top}+\mg^\top\mv=\mg^\top\ms^{*\top}=\mz^\top$, we see that both $\ms^{*\top}+\mv$ and $\ms^{*\top}$ are the solutions to $\mz^\top = \mg^\top \ms^\top$. By the uniqueness of the solution, it follows that $\ms^{*\top}+\mv=\ms^{*\top}$, which implies $\mv=\mathbf{O}_{m\times k}$. Therefore, $\Null(\mg^\top)=\{\mathbf{0}\}$.

Let $\mg^\top=(\vg_1^\top, \vg_2^\top, \cdots, \vg_m^\top)$. For $\forall \boldsymbol{x}\in\real^{m}$ such that $\mg^\top\boldsymbol{x}=\mathbf{0}$, we have $\sum_{i=1}^m\vg_i^\top x_i=\mathbf{0}$, where $x_i$ is the $i$th component of the vector $\boldsymbol{x}$. We know that $\Null(\mg^\top)=\{\mathbf{0}\}$ works if and only if $x_i=0$ for $i=1,2,\cdots,m$, which is equivalent to saying that $\vg_1^\top, \vg_2^\top,\cdots, \vg_m^\top$ are linearly independent. In other words, $\mg^\top$ has full column rank, meaning $k\geq m=\rank(\mg^\top)$. However, since $\mg^\top$ is not full column rank (as $\mg$ is not full row rank), we conclude $\ms^\top\ $(or $\ms$) cannot be uniquely recovered from $\mz=\ms\mg$ when $\mz$ and $\mg$ are known.

\subsection{Derivation of Proposition 2}\label{proof2}
\noindent\textbf{Normal Distribution}:
The likelihood of local user $i$ with $J_i$ assets is\\
$\begin{aligned}\mathcal{L}_i(y_{i1},\cdots,y_{iJ_i};\vbeta,\sigma)&=\prod_{j=1}^{J_i}\frac{1}{\sigma\sqrt{2\pi}}e^{-\frac{(y_{ij}-\combn)^2}{2\sigma^2}}\end{aligned}$. Let $\tsigma=\sigma^{-1}, \tbeta=\vbeta\tsigma$, we have the negative log-likelihood be $\ell_i^{\prime}\sim -J_i\ln\tsigma\tsigma+\sum_{j=1}^{J_i}\frac{1}{2}(\tsigma y_{ij}-\tcombn)^2$. Taking the gradient with respect to $\tsigma$ and $\tbeta$ gives $\frac{\partial \lp}{\partial \tsigma}
=\frac{-J_i}{\tsigma}+\sum_{j=1}^{J_i}(\tsigma y_{ij}-\tcombn)y_{ij},$ and $\frac{\partial \lp}{\partial \tbeta}=-\sum_{j=1}^{J_i}(\tsigma y_{ij}-\tcombn)\boldsymbol{x}_{ij}.$ 
The second order Hessian matrix is\\$\begin{aligned}\mh_i&=\begin{pmatrix}\begin{aligned}\frac{\partial \ell_i^{\prime2}}{\partial \tsigma^2}  \ \ \ \ & \frac{\partial \ell_i^{\prime2}}{\partial \tbeta^\top\partial \tsigma}\\ \frac{\partial \ell_i^{\prime2}}{\partial \tsigma\partial \tbeta} \ \ \ \  & \frac{\partial \ell_i^{\prime2}}{\partial \tbeta^\top\partial \tbeta}
\end{aligned}\end{pmatrix} = \begin{pmatrix}\begin{aligned}\frac{J_i}{\tsigma^2}+\sum_{j=1}^{J_i}y_{ij}^2 \ \ \  & -\sum_{j=1}^{J_i}y_{ij}\tx^{\top} \\ -\sum_{j=1}^{J_i}y_{ij}\tx \ \ \ & \sum_{j=1}^{J_i}\tx\tx^{\top}
\end{aligned}\end{pmatrix}. \end{aligned}
$\\
For $\forall\vb=(b_0,\vb_1^{\top})^{\top}\in\real^{K+1}$,\\ 
$\begin{aligned}\vb^\top\mh_i\vb=&\frac{J_ib_0^2}{\tsigma^2}+\sum_{j=1}^{J_i}y_{ij}^2b_0^2-\vb_1^{\top}\sum_{j=1}^{J_i}y_{ij}\boldsymbol{x}_{ij}b_0-\sum_{j=1}^{J_i}y_{ij}\boldsymbol{x}_{ij}^{\top}b_0\vb_1+\vb_1^{\top}\sum_{j=1}^{J_i}\tx\tx^{\top}\vb_1\\&=\frac{J_ib_0^2}{\tsigma^2}+\sum_{j=1}^{J_i}(\tx^{\top} \vb_1-y_{ij} b_0)^2\geq0.\end{aligned}$ 
\\
So $\mh_i$ is positive semidefinte over $\real^{K+1}$ thus guarantees the convexity of $\ell_i^{\prime}$.
\\ 

\noindent \textbf{SEV Distribution}:
The likelihood of local user $i$ with $J_i$ assets is\\$\begin{aligned}\mathcal{L}_i(y_{i1},\cdots,y_{iJ_i};\vbeta,\sigma)&=\prod_{j=1}^{J_i}\frac{1}{\sigma}e^{\frac{y_{ij}-\combn}{\sigma}}e^{-e^{\frac{y_{ij}-\combn}{\sigma}}}\end{aligned}$. Let $\tsigma=\sigma^{-1}, \tbeta=\vbeta\tsigma$, we have the negative log-likelihood be
$\ell_i^{\prime}\sim -J_i\ln\tsigma-\sum_{j=1}^{J_i}(y_{ij}\tsigma-\tcombn)+\sum_{j=1}^{J_i}e^{y_{ij}\tsigma-\tcombn}$. 
Taking the gradient with respect to $\tsigma$ and $\tbeta$ gives\\
$\begin{aligned}\frac{\partial \lp}{\partial \tsigma}
=\frac{-J_i}{\tsigma}-\sum_{j=1}^{J_i}y_{ij}+\sum_{j=1}^{J_i}\Big\{(e^{y_{ij}\tsigma-\tcombn})y_{ij}\Big\}=\frac{-J_i}{\tsigma}-\sum_{j=1}^{J_i}\Big\{(1-e^{y_{ij}\tsigma-\tcombn})y_{ij}\Big\},\end{aligned}$ 
\\
$\begin{aligned}
    \frac{\partial \lp}{\partial \tbeta}
=\sum_{j=1}^{J_i}\tx\sum_{j=1}^{J_i}\Big\{(e^{y_{ij}\tsigma-\tcombn})\tx\Big\}
=\sum_{j=1}^{J_i}\Big\{\Big[1-e^{y_{ij}\tsigma-\tcombn}\Big] \tx \Big\}.\end{aligned}$\\
The second order Hessian matrix is\\ $\mh_i=\begin{pmatrix}\begin{aligned}\frac{J_i}{\tsigma^2}+\sum_{j=1}^{J_i} y_{ij}^2 e^{y_{ij}\tsigma-\tx^{\top}\tbeta} \ \ \ & -\sum_{j=1}^{J_i}y_{ij} e^{y_{ij}\tsigma-\tx^{\top}\tbeta}\tx^{\top}\\ -\sum_{j=1}^{J_i}e^{y_{ij}\tsigma-\tx^{\top}\tbeta}y_{ij}\tx\ \ \ & \sum_{j=1}^{J_i}\tx e^{y_{ij}\tsigma-\tx^{\top}\tbeta}\tx^{\top}
\end{aligned}
\end{pmatrix}$.
For $\forall\vb=(b_0,\vb_1^{\top})^{\top}\in\real^{K+1}$, $\begin{aligned}\vb^\top\mh_i\vb=&\frac{J_ib_0^2}{\tsigma^2}+\sum_{j=1}^{J_i} b_0^2y_{ij}^2e^{y_{ij}\tsigma-\tx^{\top}\tbeta}-
2b_0\sum_{j=1}^{J_i}y_{ij}e^{y_{ij}\tsigma-\tx^{\top}\tbeta}\vb_1^{\top}\tx+
\vb_1^{\top}\Big[\sum_{j=1}^{J_i}\tx e^{y_{ij}\tsigma-\tx^{\top}\tbeta}\tx^{\top}\Big]\vb_1
\\=&\frac{J_ib_0^2}{\tsigma^2}+\sum_{j=1}^{J_i}\Big[\Big(b_0y_{ij}-\tx^{\top}\vb_1\Big)\Big(e^{y_{ij}\tsigma-\tx^{\top}\tbeta}\Big)^{\frac{1}{2}}\Big]^2
\geq0.\end{aligned}$ \\
So $\mh_i$ is positive semidefinite over $\real^{K+1}$ thus guarantees the convexity of $\ell_i^{\prime}$.
\\

\noindent \textbf{Logistic Distribution}:
The likelihood of local user $i$ with $J_i$ assets is\\$\begin{aligned}\mathcal{L}_i(y_{i1},\cdots,y_{iJ_i};\vbeta,\sigma)=\prod_{j=1}^{J_i}\frac{e^{\frac{y_{ij}-\combn}{\sigma}}}{\sigma(1+e^{\frac{y_{ij}-\combn}{\sigma}})^2}=e^{\sum_{j=1}^{J_i}\frac{y_{ij}-\combn}{\sigma}}\sigma^{-J_i}\prod_{j=1}^{J_i}(1+e^{\frac{y_{ij}-\combn}{\sigma}})^{-2}.\end{aligned}$\\
By letting $\tsigma=\sigma^{-1}, \tbeta=\vbeta\tsigma$, we have the negative log-likelihood be\\ $\ell_i^{\prime}=-\sum_{j=1}^{J_i}(y_{ij}\tsigma-\tcombn)-J_i\ln\tsigma+2\sum_{j=1}^{J_i}\ln(1+e^{y_{ij}\tsigma-\tcombn})$. 
Taking the gradient with respect to $\tsigma$ and $\tbeta$ gives\\
$
\begin{aligned}
    \frac{\partial \lp}{\partial \tsigma}=
-\sum_{j=1}^{J_i}y_{ij}-\frac{J_i}{\tsigma}+2\sum_{j=1}^{J_i}\frac{y_{ij}e^{\tsigma y_{ij}-\tcombn}}{1+e^{\tsigma y_{ij}-\tcombn}}
=\sum_{j=1}^{J_i}\Bigg\{y_{ij}\Bigg[\frac{2e^{y_{ij}\tsigma-\tcombn}}{1+e^{y_{ij}\tsigma-\tcombn}}-1 \Bigg]  \Bigg\}-\frac{J_i}{\tsigma}.\end{aligned}\\
\begin{aligned}
    \frac{\partial \lp}{\partial \tbeta}= \sum_{j=1}^{J_i}\tx-2\sum_{j=1}^{J_i}\frac{e^{y_{ij}\tsigma-\tcombn}}{1+e^{y_{ij}\tsigma-\tcombn}}\tx
=\sum_{j=1}^{J_i}\Bigg\{\Bigg[1-2\frac{e^{y_{ij}\tsigma-\tcombn}}{1+e^{y_{ij}\tsigma-\tcombn}} \Bigg] \tx\Bigg\}.\end{aligned}$
\\
The second order Hessian matrix is\\
$\mh_i= \begin{pmatrix}\begin{aligned}
\frac{J_i}{\tsigma^2}+2\sum_{j=1}^{J_i}\frac{y_{ij}^2e^{y_{ij}\tsigma-\tx^{\top}\tbeta}}{(1+e^{y_{ij}\tsigma-\tx^{\top}\tbeta})^2} &
-2\sum_{j=1}^{J_i}\frac{(y_{ij}e^{y_{ij}\tsigma-\tx^{\top}\tbeta})\tx^{\top}}{(1+e^{y_{ij}\tsigma-\tx^{\top}\tbeta})^2}
\\
-2\sum_{j=1}^{J_i}\frac{(y_{ij}e^{y_{ij}\tsigma-\tx^{\top}\tbeta})\tx}{(1+e^{y_{ij}\tsigma-\tx^{\top}\tbeta})^2} &
2\sum_{j=1}^{J_i}\frac{\tx e^{y_{ij}\tsigma-\tx^{\top}\tbeta}\tx^{\top}}{(1+e^{y_{ij}\tsigma-\tx^{\top}\tbeta})^2}
\end{aligned}
\end{pmatrix}$.\\
For $\forall\vb=(b_0,\vb_1^{\top})^{\top}\in\real^{K+1}$, \\$\begin{aligned}\vb^\top\mh_i\vb=&\frac{J_ib_0^2}{\tsigma^2}+2b_0^2\sum_{j=1}^{J_i}\frac{y_{ij}^2e^{y_{ij}\tsigma-\tx^{\top}\tbeta}}{(1+e^{y_{ij}\tsigma-\tx^{\top}\tbeta})^2}-
2\vb_1^{\top}\sum_{j=1}^{J_i}\frac{y_{ij}e^{y_{ij}\tsigma-\tx^{\top}\tbeta}\tx}{(1+e^{y_{ij}\tsigma-\tx^{\top}\tbeta})^2}b_0-
\\&2b_0\sum_{j=1}^{J_i}\frac{y_{ij}e^{y_{ij}\tsigma-\tx^{\top}\tbeta}\tx^{\top}}{(1+e^{y_{ij}\tsigma-\tx^{\top}\tbeta})^2}\vb_1+
2\vb_1^{\top}\sum_{j=1}^{J_i}\frac{(e^{y_{ij}\tsigma-\tx^{\top}\tbeta})\tx\tx^{\top}}{(1+e^{y_{ij}\tsigma-\tx^{\top}\tbeta})^2}\vb_1\\
=&\frac{J_ib_0^2}{\tsigma^2}+2\sum_{j=1}^{J_i}\Big[\Big(\frac{b_0y_{ij}-\tx^{\top}\vb_1}{1+e^{y_{ij}\tsigma-\tx^{\top}\tbeta}}\Big)\Big(e^{y_{ij}\tsigma-\tx^{\top}\tbeta} \Big)^{\frac{1}{2}} \Big]^2
\geq0.\end{aligned}$\\
So $\mh_i$ is positive semidefinite over $\real^{K+1}$ thus guarantees the convexity of $\ell_i^{\prime}$.

\bibliographystyle{ieeetr}  
\bibliography{M335}

\end{document}